\documentclass[manuscript]{acmart}

\AtBeginDocument{%
  \providecommand\BibTeX{{%
    \normalfont B\kern-0.5em{\scshape i\kern-0.25em b}\kern-0.8em\TeX}}}

\setcopyright{acmcopyright}
\copyrightyear{2018}
\acmYear{2018}
\acmDOI{XXXXXXX.XXXXXXX}

\acmConference[Conference acronym 'XX]{Make sure to enter the correct
  conference title from your rights confirmation emai}{June 03--05,
  2018}{Woodstock, NY}
\acmPrice{15.00}
\acmISBN{978-1-4503-XXXX-X/18/06}

\PassOptionsToPackage{dvipsnames,prologue}{xcolor}

\usepackage{caption}
\usepackage{subcaption}
\usepackage{type1cm}
\usepackage{lettrine}
\usepackage{amsmath}
\usepackage{listings}
\usepackage{footnote}
\usepackage[dvipsnames]{xcolor}
\usepackage{multirow}
\usepackage{svg}
\usepackage{tabularx}
\usepackage{doi}

\definecolor{codegreen}{rgb}{0,0.6,0}
\definecolor{codegray}{rgb}{0.5,0.5,0.5}
\definecolor{codepurple}{rgb}{0.58,0,0.82}
\definecolor{backcolour}{rgb}{0.95,0.97,0.97}

\lstdefinestyle{mystyle}{
    backgroundcolor=\color{backcolour},   
    commentstyle=\color{codegreen},
    keywordstyle=\color{blue},
    numberstyle=\tiny\color{codegray},
    stringstyle=\color{codepurple},
    basicstyle=\fontsize{8}{9}\selectfont\ttfamily,
    frame = single,
    breakatwhitespace=false,
    breaklines=true,
    captionpos=t,                   
    keepspaces=true,            
    showspaces=false,               
    showstringspaces=false,
    showtabs=false,                 
    tabsize=4
}

\usepackage{dblfloatfix}

\usepackage[linesnumbered,ruled]{algorithm2e}
\usepackage{amsmath, algpseudocode}
\usepackage{listings}

\makeatletter
\renewcommand{\Function}[2]{%
  \csname ALG@cmd@\ALG@L @Function\endcsname{#1}{#2}%
  \def\jayden@currentfunction{#1}%
}
\newcommand{\funclabel}[1]{%
  \@bsphack
  \protected@write\@auxout{}{%
    \string\newlabel{#1}{{\jayden@currentfunction}{\thepage}}%
  }%
  \@esphack
}
\makeatother

\definecolor{notecolor}{rgb}{0.75,0,0} 

\definecolor{darkgreen}{RGB}{5, 150, 73}

\newcommand{\shortener}{\textsc{SkipShortener}\xspace}
\newcommand{\remover}{\textsc{SkipRemover}\xspace}

\newcommand{\tool}{\textsc{Tailor}\xspace}

\usepackage{hyperref}

\usepackage{ifthen}
\usepackage[normalem]{ulem} 
\usepackage{xcolor}

\newboolean{showedits}
\setboolean{showedits}{true} 
\ifthenelse{\boolean{showedits}}
{

	\newcommand{\del}[1]{\textcolor{red}{\sout{#1}}}
	
}{

	\newcommand{\del}[1]{}
	
}

\newboolean{showcolors}
\setboolean{showcolors}{true} 
\ifthenelse{\boolean{showcolors}}
{

}{

}

\newboolean{showcomments}
\setboolean{showcomments}{true}

\ifthenelse{\boolean{showcomments}}
{\newcommand{\nbc}[3]{
 {\colorbox{#3}{\bfseries\sffamily\scriptsize
 	\textcolor{white}{#1}}}
 {\textcolor{#3}{\sf\small$\blacktriangleright
 	${#2}$\blacktriangleleft$}}}
 }
{\newcommand{\nbc}[3]{}}

\definecolor{ibcolor}{rgb}{0.4,0.6,0.2}

\definecolor{metacolor}{rgb}{0.5,0,0.5}

\definecolor{ideacolor}{rgb}{1.0,0,0.5}

\definecolor{todocolor}{rgb}{0.9,0.1,0.1}

\definecolor{qcolor}{rgb}{0.2,0.0,0.9}

\definecolor{owcolor}{rgb}{0.2,0.4,0.2}

\definecolor{rkcolor}{RGB}{55, 109, 160}

\definecolor{amcolor}{RGB}{55, 109, 160}

\definecolor{amcolor}{RGB}{0,206,209}

\definecolor{agcolor}{RGB}{111, 23, 189}

\begin{document}

\title{\tool: Altering Skip Connections for Resource-Efficient Inference}

\author{Olivia Weng}
\email{oweng@ucsd.edu}
\author{Gabriel Marcano}
\affiliation{%
  \institution{University of California San Diego}
  \country{USA}
}

\author{Vladimir Loncar}
\affiliation{%
  \institution{Massachusetts Institute of Technology}
  \country{USA}}

\author{Alireza Khodamoradi}
\affiliation{%
  \institution{AMD}
  \country{USA}
}

\author{Abarajithan G}
\author{Nojan Sheybani}
\affiliation{%
 \institution{University of California San Diego}
 \country{USA}}

\author{Andres Meza}
\author{Farinaz Koushanfar}
\affiliation{%
 \institution{University of California San Diego}
 \country{USA}}

\author{Kristof Denolf}
\affiliation{%
  \institution{AMD}
  \country{USA}}

\author{Javier Mauricio Duarte}
\author{Ryan Kastner}
\affiliation{%
  \institution{University of California San Diego}
  \country{USA}}

\renewcommand{\shortauthors}{Weng, et al.}

\begin{abstract}
 Deep neural networks use skip connections 
to improve training convergence. 
However, these skip connections are costly in hardware, requiring extra buffers and increasing on- and off-chip memory utilization and bandwidth requirements.
In this paper, we show that skip connections can be optimized for hardware when tackled with a hardware-software codesign approach.
We argue that while a network's skip connections are needed for the network to learn, they can later be removed or shortened to provide a more hardware efficient implementation with minimal to no accuracy loss.
We introduce \tool, a codesign tool whose hardware-aware training algorithm gradually removes or shortens a fully trained network's skip connections to lower their hardware cost. 
\tool improves resource utilization by up to 34\% for BRAMs, 13\% for FFs, and 16\% for LUTs for on-chip,
dataflow-style architectures.
\tool increases performance by 30\% and reduces memory bandwidth by 45\% for a 2D processing element array architecture.

\end{abstract}

\ccsdesc[500]{Hardware~Hardware-software codesign}
\ccsdesc[500]{Computer systems organization~Neural networks}

\keywords{Hardware-software co-design, neural networks}

\received{20 February 2007}
\received[revised]{12 March 2009}
\received[accepted]{5 June 2009}

\maketitle

\section{Introduction}\label{sec:introduction}
Convolutional neural networks (NNs) often rely on skip connections---identity functions that combine the outputs of different layers---to improve training convergence~\cite{he_deep_2015, veit2016residual}. 
Skip connections help mitigate the vanishing gradient problem~\cite{bengio_learning_1994, glorot2010understanding} that occurs when training large CNNs, which helps increase the network's accuracy. 
Skip connections allow NNs to have fewer filters/weights than architectures that lack skip connections~\cite{he_deep_2015}, such as VGG~\cite{vgg2015}.

\begin{figure}[t]
\centering
\subfloat[Traditional]{\label{fig:resblock}
    \includegraphics[width=0.15\textwidth]{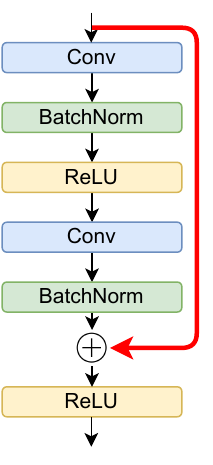}}
    \hspace{10mm}
\subfloat[Shortened]{\label{fig:shortresblock}
    \includegraphics[width=0.15\textwidth]{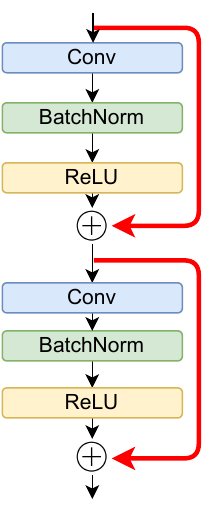}}
    \hspace{10mm}
\subfloat[None]{\label{fig:nonresblock}
    \includegraphics[width=0.15\textwidth]{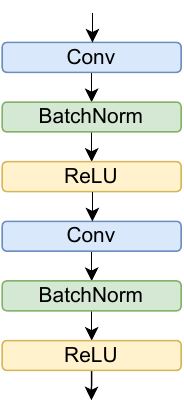}}
    
    \caption{Neural networks with traditional skip connections, like ResNet (a), have inefficient hardware implementations because the skip connection data must be preserved in memory during five layers of computation. 
    This irregular topology increases memory resources and bandwidth. 
    A more regular topology with reduced skip connection lifetimes would use fewer resources. 
    \tool achieves this by shortening skip connections (b) or by eliminating them completely (c). Skip connections are in red.}  
\end{figure}

However, skip connections are generally detrimental to hardware efficiency. 
They have an irregular design that is ill-suited for hardware acceleration. 
This is due to their long lifetimes, which span several NN layers, increasing memory utilization and bandwidth requirements.
This is particularly true in ResNets~\cite{he_deep_2015}, which introduced skip connections that spanned across five layers: two convolutions, two batch normalizations (BNs), and a ReLU activation~\cite{relu1,relu2} (see \autoref{fig:resblock}). %
The skip connection involves minimal computation---it is either the identity or a 1$\times$1 convolutional layer for scaling---but it extends the necessary lifespan of the input data. 
Thus, we must store skip connection data for the duration of time needed to compute the five NN layers.
In total, a model's skip connection data accounts for $\sim$10\% of its memory bandwidth either on or off chip. 
Buffering skip connections on chip increases on-chip memory utilization, whereas moving them off chip not only increases off-chip memory bandwidth but also requires extra control logic for scheduling~\cite{controllogic17, offchipmemory}.





Optimizing skip connections requires careful \textit{hardware-software codesign}.
Skip connections are crucial for model convergence; naively removing them to reduce hardware resources leads to low accuracy~\cite{diracnets17, avoidingskip18}.
Instead, we must codesign how the model is (1) trained and (2) implemented in hardware to achieve a model that is both accurate and resource-efficient.

We develop \tool, a codesign method that gradually alters a NN's skip connections during training to produce a highly accurate and resource-efficient NN.  
Our results in \autoref{sec:results} show that \tool can remove or shorten skip connections to achieve topologically regular NNs~(\autoref{fig:shortresblock} and \ref{fig:nonresblock}) that substantially reduce hardware resources, reduce memory bandwidth, and increase performance with minimal to no accuracy loss.

\tool takes an \textit{existing} pre-trained model and reduces the hardware complexity of its skip connections with minimal to no accuracy loss. 
Moreover, \tool exploits the flexiblity of the FPGA architecture to customize the skip connection memories, which is not possible on a GPU or CPU.
\tool accomplishes this dynamically during retraining in one of two ways: (1) \remover removes the skip connections altogether (\autoref{fig:nonresblock}) to eliminate all associated hardware complications or (2) \shortener shortens each skip connection by splitting it into multiple shorter ones (\autoref{fig:shortresblock}).

We evaluate \tool's applicability and benefit on ResNets~\cite{he_deep_2015, he_resnet_2016} and QuartzNets~\cite{kriman2020quartznet}---two important classes of NNs that contain skip connections of varying lengths.
We also study implementing skip connections with an on-chip, dataflow-style FPGA architecture using hls4ml~\cite{aarrestad2021fast, fahim2021hls4ml} and a 2D array of multiply-accumulate processing elements.
\tool reduces resource utilization of hls4ml architectures by up to 34\% for BRAMs, 13\% for FFs, and 16\% for LUTs.  
\tool increases the performance of 2D array architecture by 30\% and reduces memory bandwidth by 45\%.. 

{\tolerance=7000
\tool's hardware-software codesign approach reduces hardware complexity and resources by altering skip connections dynamically during retraining. 
Our contributions are:
\begin{itemize}
    \item the \tool software methodology of removing or shortening skip connections from existing NNs with minimal to no loss in accuracy,
    \item the \tool hardware designs that exploit FPGA-specific architecture optimizations, which are not possible on GPU/CPU, to produce less resource-intensive skip connection implementations,
    \item experiments demonstrating that \shortener and \remover models are implemented more efficiently with better performance and resource utilization than their traditional skip connection counterparts,  
    \item and public release of the Tailor hardware-software codesign framework~\cite{Tailor}.
\end{itemize}
\par}

In \autoref{sec:background}, we review related work. 
In \autoref{sec:tool}, we explain how \tool's NN alterations optimize the hardware architecture. 
We then describe \tool's two training methods, \remover and \shortener, that alter skip connections with little to no loss in accuracy. 
\autoref{sec:results} provides training, quantization, and hardware results for \remover and \shortener. 
\autoref{sec:discussion} discusses the tradeoffs \tool presents between accuracy and hardware resource reductions. 
\autoref{sec:conclusion} concludes the paper.

\section{Background}
\label{sec:background}
\subsection{Removing Skip Connections}
\label{sec:removeskipbackground}
While skip connection removal has been studied before~\cite{diracnets17, zagoruyko_diracnets_2018, avoidingskip18, ding2021repvgg, li2020residual}, prior work is lacking in several ways: (1) preliminary work~\cite{diracnets17, zagoruyko_diracnets_2018, avoidingskip18} only studies shallow models (up to 34 layers); (2) Li et al.~\cite{li2020residual} do not remove all of the skip connections in the models they evaluate; (3) Ding et al.~\cite{ding2021repvgg} and Li et al.~\cite{li2020residual} both have limited architectural evaluations (e.g., GPU \& mobile) that do not consider the highly customized skip connections memories enabled by FPGAs; and (4) Ding et al.~\cite{ding2021repvgg} require starting with an entirely new NN topology whose skip connections are removable.

Monti et al.~\cite{avoidingskip18} start with a standard ResNet and introduce a new training method. 
This method uses an objective function that penalizes the skip connections and phases them out by the end of the training. 
This technique has only been applied to smaller ResNets (18 to 34 layers) with a small decrease in accuracy between 0.5 and 3\%.

Zagoruyko and Komodakis~\cite{diracnets17} also develop a method for removing skip connections in a NN. 
They replace skip connections with Dirac parameterization, creating a new NN called DiracNet. 
The Dirac parameterization is shown in \autoref{eq:diracnet},
\begin{align}
    \text{DiracNet~\cite{diracnets17}: } y &= \sigma(x+Wx)\label{eq:diracnet} \\
    \text{ResNet~\cite{he_deep_2015}: } y &= x+\sigma(Wx)\,,\label{eq:resnet}
\end{align}
where $\sigma(\cdot)$ is the nonlinear activation function, $W$ is the layer weight matrix, $x$ is the layer input, and $y$ is the layer output.
For ease of comparison with ResNets, \autoref{eq:resnet} is simplified to show only one convolutional layer. 
In fact, skip connections in ResNets hop over more than one convolutional layer, while in DiracNets, the identity mapping is over one single convolutional layer. 
Therefore, the weights and the identity mapping of the input can be folded because $x+Wx=(I+W)x$. 
This change requires DiracNets to widen the NN layers in the ResNets that they started with. 
The authors showed that their technique could be used to create models with up to 34 layers. 
Although it works for shallower models, DiracNets show a decrease in accuracy between 0.5\% and 1.5\% compared to ResNets. 
In contrast, \remover eliminates skip connections without widening the layers in the NN and does not need to make this accuracy tradeoff.

Li et al.~\cite{li2020residual} develop residual distillation (RD), which is a modified knowledge distillation framework. 
RD starts with a ResNet as the teacher and a plain CNN without skip connections as the student. 
Unlike standard knowledge distillation, RD passes the teacher's gradients to the student during training. 
This differs from \tool because RD starts with a student model without skip connections, whereas \tool \textit{gradually} modifies a model's skip connections every few epochs during training without sharing gradients. 
Moreover, while RD removes all skip connections from models evaluated on simpler datasets like CIFAR-10 and CIFAR-100~\cite{cifardataset}, it fails to remove all skip connections in its ImageNet evaluation, leaving 18\% of them in the network, which is a costly choice. 
In our ImageNet evaluation (see \autoref{sec:training}), our \remover method removes all skip connections with minimal accuracy loss. 

Ding et al.~\cite{ding2021repvgg} introduce a new model architecture RepVGG, which trains using 3$\times$3 convolutional layers that are each skipped over by both a 1$\times$1 convolution and an identity connection. 
At inference time, these connections can be re-parameterized into the parameters of the 3$\times$3 convolutional layers. 
While RepVGG is more accurate than our \remover model, it requires starting from their specialized training model architecture. 
This is costly to developers who have already trained a model with skip connections on their dataset.
Similarly, transferring a pre-trained RepVGG model to a new dataset via transfer learning can be time-consuming given the many different methods~\cite{weiss2016survey, zhuang2020comprehensive, pan2009survey} to evaluate.
As such, \tool is ideal for these developers because it modifies the skip connections of an \textit{existing} pre-trained model to be more resource-efficient with minimal to no accuracy loss. 
Developers can leverage the training they have already done and need not start from scratch with a brand new RepVGG architecture.  

\subsection{Simplifying Skip Connection Hardware}
ShuffleNet~\cite{ma2018shufflenet}, DiracDeltaNet~\cite{yang2019synetgy}, and DenseNet~\cite{huang2017densely} simplify skip connections by making them \emph{concatenative}, i.e., they concatenate, rather than add, the skip connection data to the output of a layer. 
Concatenative skip connections take advantage of the fact that spatially consecutive memory accesses are typically faster than random accesses. 
This concatenation and off-chip data movement is possible using a simple controller (e.g., DMA engine). 

\tool uses two techniques to simplify the skip connection hardware. \remover eliminates all logic and memory needed for a skip connection, making them less expensive than concatenative skip connections. 
Careful retraining allows skip connection removal in smaller networks with no degradation in accuracy. For larger networks,  \shortener shortens the additive skip connections. 
By reducing their lifespans, the hardware implementation requires fewer resources. 
\shortener is not necessarily simpler than ShuffleNet~\cite{ma2018shufflenet} or DiracDeltaNet~\cite{yang2019synetgy}. 
However, these concatenative skip connections have only been evaluated on image classification and object detection tasks. 
In our work, we demonstrate our \remover and \shortener methods on multifarious NNs and classification tasks, namely image-classifying ResNets of varying depths, DNA-basecalling QuartzNet-5$\times$5, and automatic-speech-recognizing QuartzNet-10$\times$5. 
With respect to DenseNet~\cite{huang2017densely}, \shortener ResNets use much less memory and bandwidth because DenseNet relies on significantly more skip connections throughout its NN. 
Given a NN with $L$ layers, DenseNet needs the memory and bandwidth to execute $L(L + 1)/2$ concatenative skip connections, compared with \shortener ResNets' mere $L$ skip connections. 
With so many more skip connections, DenseNet is more expensive for hardware than \shortener ResNets. 

Finally, all these techniques simplify skip connection hardware from the outset, building their models with modified skip connections and then training them from scratch.
\tool differs because its hardware-aware training method \emph{dynamically} alters the skip connections every few epochs during training, taking advantage of what the NN has learned with skip connections.
Thus \tool allows the NN to gradually adapt to shortened skip connections (\shortener) or none at all (\remover).



\section{\tool}
\label{sec:tool}
Skip connections are important for training (to provide good accuracy), yet complicate implementation (requiring additional hardware resources and reducing performance).
\tool modifies skip connections to make their hardware implementation more efficient. \tool uses a retraining method that gradually alters the network, resulting in little to no loss in accuracy.

\subsection{Hardware Design}\label{sec:hw-motivation}
\begin{figure}[t]
\centering
\subfloat[Traditional skip connection architecture]{\label{fig:trad-arch}
    \includegraphics[width=0.9\textwidth]{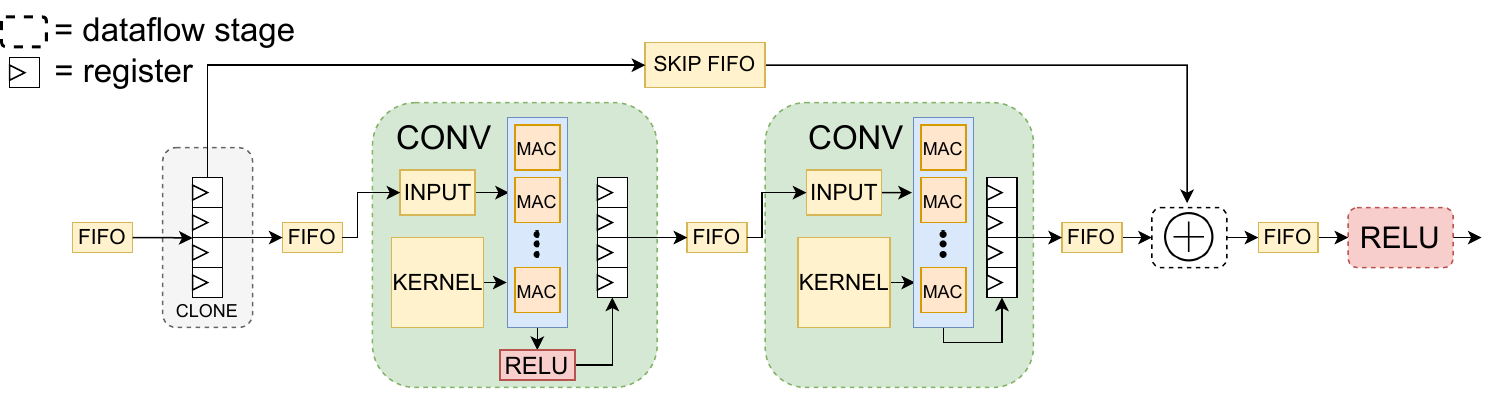}}
    \vspace{4mm}
\subfloat[No skip connection architecture]{\label{fig:no-skip-arch}
    \includegraphics[width=0.48\textwidth]{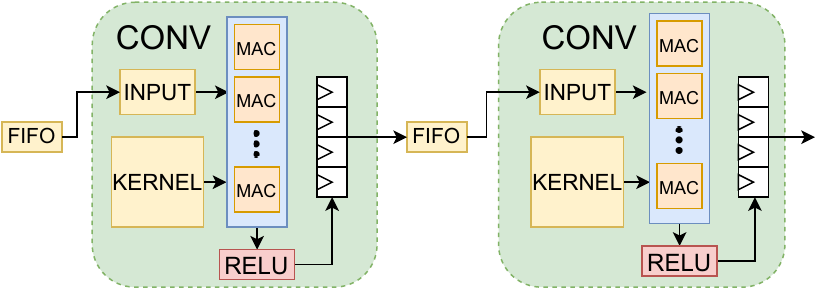}}
    \vspace{4mm}
\subfloat[Shortened skip architecture]{\label{fig:short-arch}
    \includegraphics[width=0.48\textwidth]{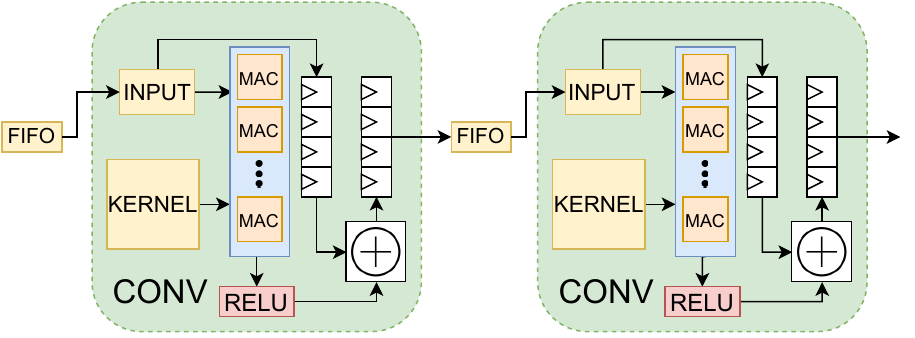}}
    \caption{The hls4ml hardware architectures for traditional, shortened, and no skip connections. 
hls4ml pipelines each layer as is common for latency-critical tasks in resource-constrained environments~\cite{aarrestad2021fast, fahim2021hls4ml}.
The three architectures correspond to a ResNet implemented with a traditional skip connection (a), shortened skip connections (b), and no skip connections (c). 
Note that we combine the batch normalization parameters with the kernel, as is commonly done~\cite{jacob2018quantization}.
}
    \label{fig:arch}
\end{figure}

\autoref{fig:arch} shows three hardware implementations for NNs with traditional, shortened, and no-skip connections. 
The implementations correspond to accelerators produced by hls4ml---a tool that translates Python models into high-level synthesis code~\cite{duarte2018fast}. 
hls4ml creates a separate datapath for each layer and performs task-level pipelining across the layers.
The layers communicate using FIFOs (AXI streams).
Everything encapsulated by a dashed line resides in one pipeline stage.
The inputs are fed into the architecture using a stream, and the results are given as an output stream. 
The weights are all stored on-chip, and all the internal results are stored on-chip.
We evaluate each of these designs on FPGA later in \autoref{sec:hw-results} along with another style of architecture using a 2D array of processing elements. 
\tool allows us to trade off between accuracy, performance, and resource usage through co-design of the neural network using hardware-aware training.

\autoref{fig:trad-arch} shows the hardware needed to implement a single ResNet's skip connection. 
Note that in all of the designs shown in \autoref{fig:arch}, we fuse the batch normalization parameters with the kernel, as is commonly done~\cite{jacob2018quantization}.
To be low latency and high throughput, the design uses task-level pipelining (i.e., the HLS dataflow pragma) for each NN layer, or a small grouping of layers, and streams the data between each dataflow stage using first-in first-out buffers (FIFOs).
Since FIFOs can only be read from once, skip connections complicate the design. 
We must spend a dataflow stage on cloning the skip connection data from its input FIFO into two other FIFOs so that it can be read twice for its two datapaths. 
The first path goes through a collection of convolutional and ReLU layers, and the second stores the data in a FIFO exclusive to skip connections (skip FIFO). 
Once the data has gone through the first path, we read from the skip FIFO to perform the addition to complete the skip connection's identity function. 
As such, implementing a skip connection on chip requires several extra FIFOs for handling the skip connection data, and this in turn increases on-chip memory resource utilization. 

Ideally, we would eliminate the skip connections. 
As seen in \autoref{fig:no-skip-arch}, without skip connections, we cut the number of dataflow stages in half (no more Clone, Add, or ReLU stages) and use less than half of the requisite FIFOs compared with \autoref{fig:trad-arch}. 
All we need to do is pass the data through the convolutional and ReLU layers. 
This reduces resource utilization by up to 16\% (see \autoref{sec:hw-results}).

It may not be possible to remove the skip connections because they are essential for training convergence. 
In these cases, shortening the skip connections can simplify their hardware implementation. 
\autoref{fig:short-arch} shows a modified network with shortened skip connections such that \emph{each skip connection's lifespan resides within a single dataflow stage}.
We do not need additional dataflow stages to clone skip connection data.
The shorter lifespans allow the shortened skip connections to be stored in \textit{shift registers}, which can be implemented using the more abundant FFs as opposed to BRAMs, which is used in the traditional skip connection's hardware design.
In this way, we exploit the short skip connections' lifetimes and use simpler, more efficient hardware memories to implement them (see \autoref{sec:hw-results}).
As such, we achieve a similar architecture to the version without skip connections (\autoref{fig:no-skip-arch}), and similarly reduce resources spent on additional dataflow stages and FIFOs in \autoref{fig:trad-arch}. 
\shortener is thus more resource-efficient than the traditional skip connection design. 
In fact, \shortener provides a tradeoff between the \remover and traditional designs because it uses more resources than \remover but less than the traditional one (see \autoref{sec:hw-results}). 
But as we later show in \autoref{sec:training}, \shortener maintains accuracy in cases where \remover accuracy drops off.
Thus, \shortener allows for design space exploration to balance accuracy and resource usage.

When used with hls4ml, \tool reduces resource consumption without changing the performance. 
This is a consequence of hls4ml's dataflow design; the resources we remove are not on the critical path---they are operating in parallel to the critical path. 
A dataflow design uses task-level pipelining, so reducing the resources spent on stages not on the critical path does not help or hurt overall throughput. 
Based on our Vivado co-simulation results, the clone stage executes in microseconds while the convolutional layer executes in milliseconds, an order of magnitude difference. 
Therefore, removing the clone buffer (\autoref{fig:no-skip-arch}) or implementing it more efficiently (\autoref{fig:short-arch}) will not affect the overall dataflow latency because its latency is an order of magnitude less than the convolution's latency. 
This means \tool's resource reductions do not increase or decrease latency or throughput for this architecture style, as later shown in \autoref{tab:latency}.

\begin{figure}[ht]
\centering
    \includegraphics[width=\textwidth]{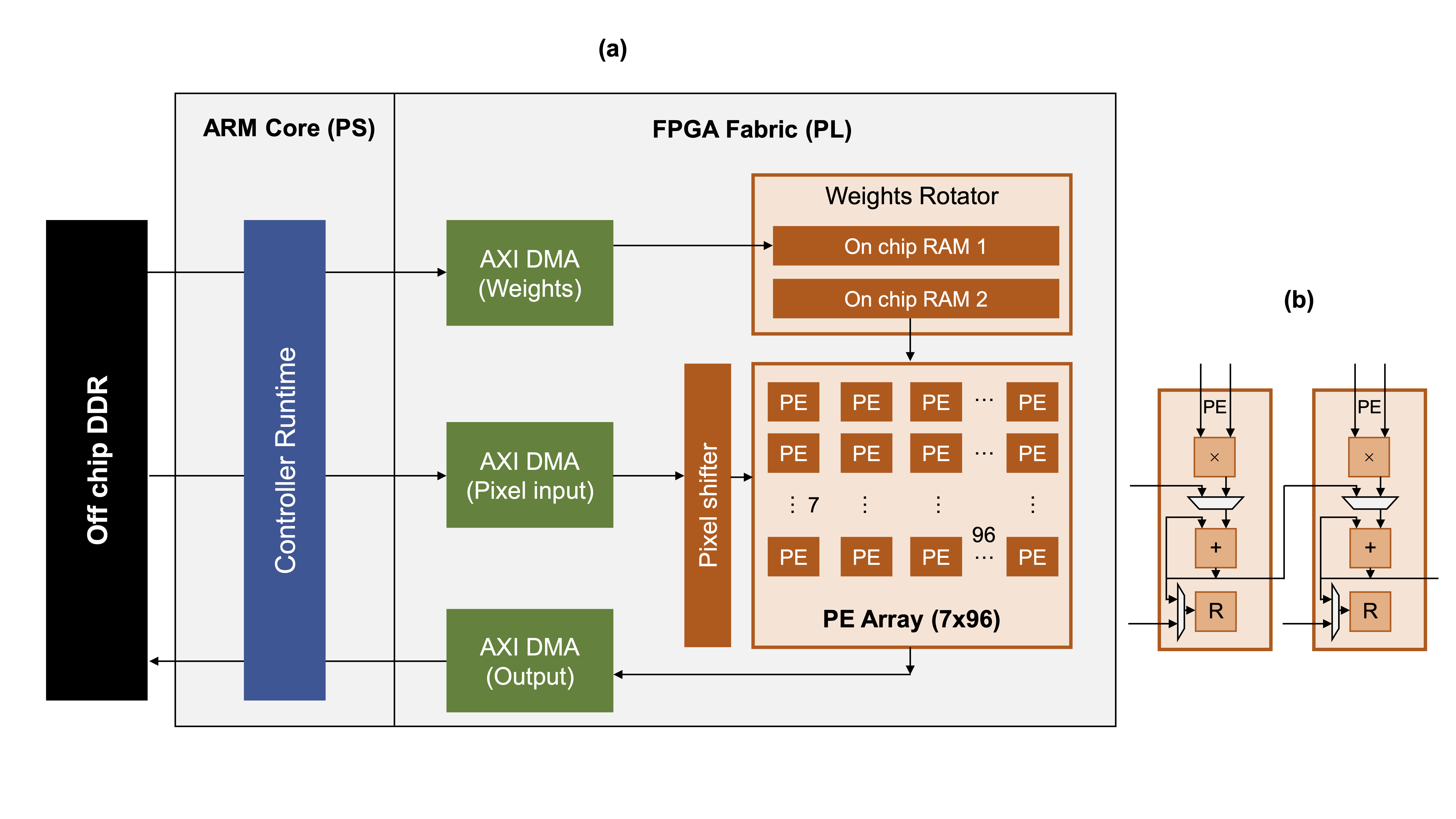}
    \caption{(a) A Reconfigurable DNN Architecture synthesized on a ZCU102 FPGA development board. The architecture has a 2D array of processing elements that are iteratively programmed to compute layer operations. The controller runtime programs the DMA engines to load off-chip inputs and weights and store the intermediate and final results off-chip. (b) The processing elements (PE) are a multiply-accumulate datapath. }
\label{fig:dnn_engine}
\end{figure}

Another prevalent style of FPGA CNN architectures instantiates a 2D processing element (PE) array and iteratively programs the convolutions and other operations onto that PE array.
We call this style of computation a Reconfigurable DNN Architecture.
\autoref{fig:dnn_engine} provides an example architecture used in our experiments.
We build this architecture using DeepSoCFlow~\footnote{\url{https://github.com/abarajithan11/deepsocflow}}.
Following the taxonomy described in \cite{juracy2023cnn}, the reconfigurable DNN architecture is a 2D array of processing elements that optimally perform standard convolution and matrix multiplication with high data reuse. 
The dataflow is primarily output stationary while prioritizing maximal weight reuse and also reusing inputs to an extent. 
The engine performs fixed-point computations, where the input, weight, and output bit widths are adjustable as synthesis parameters, along with the number of rows and columns of processing elements. 
The weights rotator prefetches the weights of the next iteration into one block of on-chip memory while the other bank delivers weights, rotating them hundreds of times for maximal data reuse. 
The Pixel Shifter shifts perform vertical convolution. 
Partial sums are shifted to the PE on the right to compute horizontal convolution. 
The results are streamed out through the output DMA to the off-chip memory. 
The runtime controller would perform the residual addition, quantization, and activation on the processing system side while the engine computes the next iteration. 
Our implementation uses the ARM processor available in the Zynq chip. 
FPGAs without processors could instantiate a softcore processor to perform the controller runtime operations.

The \tool optimizations have different effects on the Reconfigurable DNN architecture as compared to hls4ml architecture.
The Reconfigurable DNN architecture computes skip connections by loading input data from off-chip memory and performing the required operations upon it (addition, convoluation) 
Thus, unlike in hls4ml, removing a skip connection does not change the architecture; instead it changes how computations are mapped to that architecture. 
Skip connection removal eliminates the need to fetch the skip connection data and perform the associated convolution and addition operations. 
This increases the overall performance as we describe in \autoref{sec:results}.

\subsection{Hardware-aware Training}

\begin{figure*}[t]
    \centering
\subfloat[\remover]{\label{fig:remover}
    \includegraphics[width=0.8\textwidth]{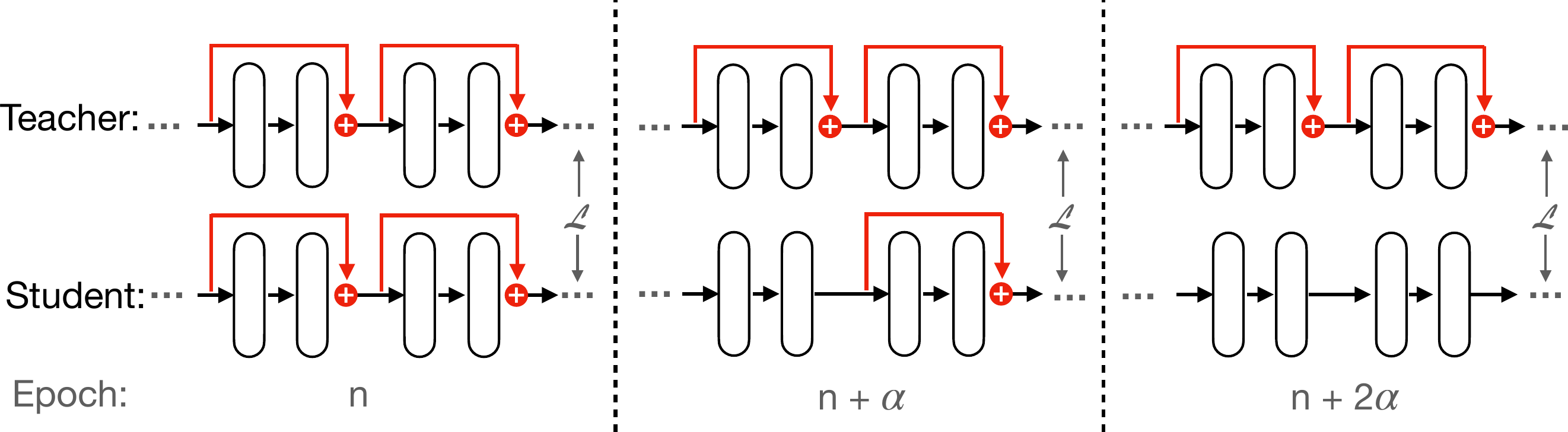}}
    \hspace{4mm}
\subfloat[\shortener]{\label{fig:shortener}
    \includegraphics[width=0.8\textwidth]{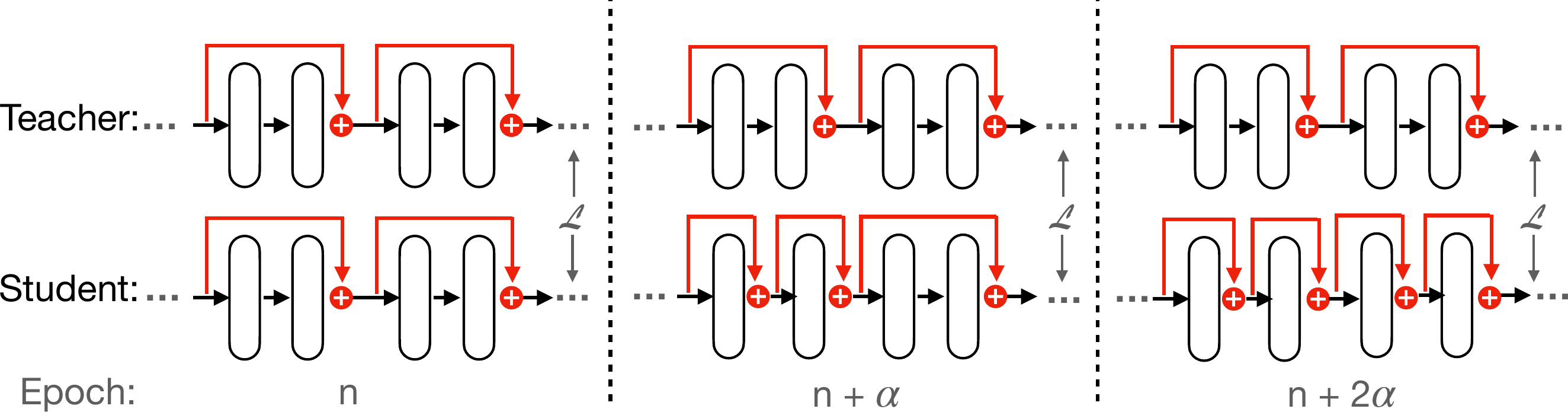}}
    \caption{Three iterations in the \remover and \shortener algorithms as applied to a ResNet. 
    In this example, skip connections are altered every $\alpha$ epochs and $\alpha|n$. 
    Each pill block represents a set of convolutional, BN, and ReLU layers, and the skip connections are in red. 
    $\mathcal{L}$ is the KD loss function defined in \autoref{eq:lossKD}. Only the student model is used for inference.}
    \label{fig:tool-alg}
\end{figure*}

It is difficult to modify a NN's skip connections without reducing accuracy. 
Naively removing all skip connections before or after training a NN is detrimental to its accuracy. 
Instead, \tool consists of two training algorithms, \emph{\remover} and \emph{\shortener}, that gradually alter a NN's skip connections on the fly---removing or shortening them every few epochs---in order to make them resource-efficient. 
Gradually altering the model during training tempers the performance drop of removing or shortening the skip connections, yielding minimal to no loss in accuracy as well as significant advantages in the hardware implementation, as described above.

\tool's iterative learning approach finetunes the altered NNs using a compression method known as \emph{knowledge distillation (KD)}~\cite{hinton_distilling_2015}.
KD distills the knowledge of a larger, more complex NN (the teacher) into a smaller, simpler NN (the student).
While the student model is training, it compares its output to the teacher model's output and thus learns from the teacher to perform better. 
KD provides impressive results for compressing NNs for various applications~\cite{Mirzadeh2020ImprovedKD, kd2016, kd2017}.
In traditional KD, the teacher model is already trained, and the student model is trained to match the teacher's behavior by replicating its output. 
The student achieves this by training with a loss function
\begin{align}
\label{eq:lossKD}
    \mathcal{L} = (1-\beta)\mathcal{G}(\ell, s) + \beta \mathcal{H}(t, s)
\end{align}
where $\mathcal{G}$ and $\mathcal{H}$ are distance functions, $s$ and $t$ are student and teacher output vectors respectively, $\ell$ is the correct label vector, and $\beta$ is a tunable parameter~\cite{hinton_distilling_2015}.

With this idea in mind, both \remover and \shortener start with two identical pre-trained NNs with traditional skip connections, where one serves as the teacher and the other serves as the student. 
During the retraining stage, \remover removes a given skip connection every few epochs.
\shortener takes a similar iterative approach and, every few epochs, splits a given skip connection into multiple shorter ones. 
The skip connections are removed or shortened starting from the first skip connection encountered in the NN (from the input) to the last.

\autoref{fig:tool-alg} visualizes both \remover's (\autoref{fig:remover}) and \shortener's (\autoref{fig:shortener}) training algorithms for a ResNet-style NN. 
During training, we remove (\remover) or shorten (\shortener) one of the student's skip connections every $\alpha$ epochs.
If $n$ is divisible by $\alpha$ (as in \autoref{fig:tool-alg}), then at epoch $n$, the student has had $n/\alpha$ skip connections altered, and we are viewing the next two skip connections to be modified in the student model: the $(n/\alpha) + 1$st and $(n/\alpha) + 2$nd.
At epoch $n+\alpha$, the $(n/\alpha) + 1$st skip connection is altered (removed under \remover or split into two shorter skip connections under \shortener).
The NN then trains for $\alpha$ epochs so that the student model can improve its weights given the latest model topology.
Afterwards, at epoch $n + 2\alpha$, the $(n/\alpha) + 2$nd skip connection is similarly altered. 
During the entire skip modification retraining process, the student uses the KD loss function $\mathcal{L}$ defined in \autoref{eq:lossKD} to learn from the teacher and the true labels. 
The teacher's model topology and weights remain fixed during training. 
Once all skip connections have been altered, the student model continues training under KD for the remaining number of training epochs as defined by the user.
Only the student model is used for inference.

\tool is novel because it \emph{dynamically} transforms skip connections every few epochs during training. 
This is an instance of \emph{hardware-aware training} because the skip connection are slowly altered specifically to reduce hardware resources, as previously discussed in \autoref{sec:hw-motivation}.
The gradual skip connection alterations allow the NN to take advantage of what it has learned with skip connections, so that it can dynamically adapt to shortened skip connections (\shortener) or none at all (\remover).
\autoref{alg:hw-aware-training} describes \tool's hardware-aware training process.

\begin{algorithm}
\DontPrintSemicolon
\caption{\textsc{Hardware-aware Training}}	
\label{alg:hw-aware-training}
set $alter$ \tcp{REMOVE or SHORTEN}
let $\alpha$ = how often to modify a skip connection\\
\textit{teacher} = pre-trained model\\
\textit{student} = pre-trained model\\
let current-skip = \textit{student}'s first skip connection from the input side \\
let current-layers = all layers skipped by current-skip \\
\SetKwFunction{Remover}{SkipRemover}
\SetKwProg{Fn}{Function}{:}{}
\Fn{\Remover{current-skip}}{ 
    \tcp{see \autoref{fig:remover}}
    remove current-skip\;
    \KwRet student model's next skip connection from the input side\;
}
\SetKwFunction{Shortener}{SkipShortener}
\SetKwProg{Fn}{Function}{:}{}
\Fn{\Shortener{current-skip, current-layers}}{ 
    \tcp{see \autoref{fig:shortener}}
    Split current-skip into $len$(current-layers) skip connections\;
    current-skip = student model's next skip connection from the input side\;
    current-layers = student's next layers skipped by the new current-skip\;
    \KwRet current-skip, current-layers\;
}

\For{$i$ in epochs}{
    \If{ $i \neq 0$ and $i \mod \alpha = 0$}{
        \uIf{$alter$ == REMOVE}{
            current-skip = \Remover{current-skip}\;
        }
        \uElseIf{$alter$ == SHORTEN}{
            current-skip, current-layers = \Shortener{current-skip, current-layers}\;
        }
    }
    train \textit{student} using \autoref{eq:lossKD}
}
save the student model
\end{algorithm}

\section{Results}
\label{sec:results}
We evaluate \tool on two popular kinds of NNs that rely on skip connections: ResNets~\cite{he_deep_2015} and  QuartzNets~\cite{kriman2020quartznet}. 
We study the effects of \tool on model accuracy, quantization, and hardware resource utilization. 

\subsection{Training results}\label{sec:training}
To evaluate how \tool affects a NN's accuracy, we train ResNets and QuartzNets of varying depths using our \remover and \shortener algorithms in PyTorch~\cite{NEURIPS2019_pytorch}. 
The ResNets range from 20 to 110 layers and are trained on the CIFAR-10~\cite{cifardataset}, CIFAR-100~\cite{cifardataset}, and SVHN~\cite{svhndataset} datasets. 
We also evaluate ResNet50, which has a different skip connection topology than standard ResNets, on the ImageNet dataset~\cite{deng2009imagenet}.
The QuartzNets span between 29 and 54 layers. 
Their structure is determined by the number and lifetimes of their skip connections. For instance, a QuartzNet-10$\times$5 has 10 skip connections that each have a lifetime of 5 sets of layers. 
We train a QuartzNet-5$\times$5 on the Oxford Nanopore Reads dataset~\cite{silvestre2021pair}, a DNA basecalling task. 
We also train a QuartzNet-10$\times$5 on the LibriSpeech dataset~\cite{panayotov2015librispeech}, an automatic speech recognition (ASR) task, which converts speech audio to text. 
ASR tasks are assessed using word error rate (WER), which measures the percent of words that the model predicted incorrectly. 
In all of our ResNet and QuartzNet-10$\times$5 training experiments, we set $\alpha = 3$ in \autoref{alg:hw-aware-training}, so skip connections are removed or shortened every three epochs. 
For QuartzNet-5$\times$5, we set $\alpha = 1$ instead because it trains better this way.  
For the ResNets, we set $\mathcal{G}$ and $\mathcal{H}$ in \autoref{eq:lossKD} to \emph{categorical cross entropy} and \emph{mean-squared error}, respectively, and set $\beta = 0.35$. 
For the QuartzNets, we set \autoref{eq:lossKD}'s parameters similarly, except for $\mathcal{G}$, which we set to \emph{connectionist temporal classification loss}, which is used to train difficult tasks involving sequence alignment (like DNA basecalling and ASR).
Note that in our training results, ``Baseline'' refers to the unmodified NN counterpart with conventional skip connections.

\begin{figure}[t]
    \centering
    \subfloat[CIFAR-10]{\label{fig:cifar10}
    \includegraphics[width=0.3\textwidth]{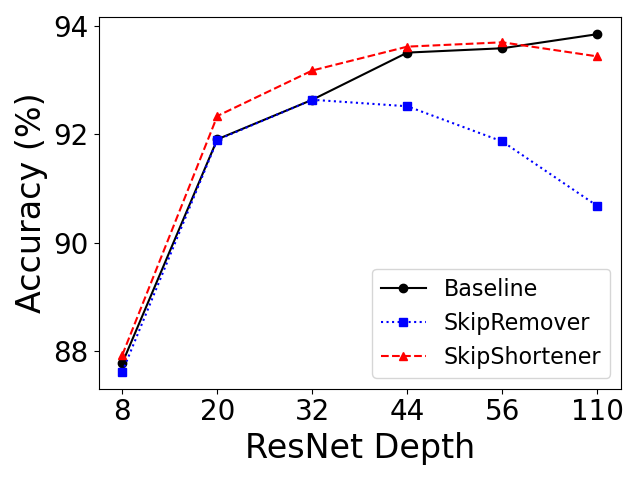}}
    \hspace{1mm}
    \subfloat[CIFAR-100]{\label{fig:cifar100}
    \includegraphics[width=0.3\textwidth]{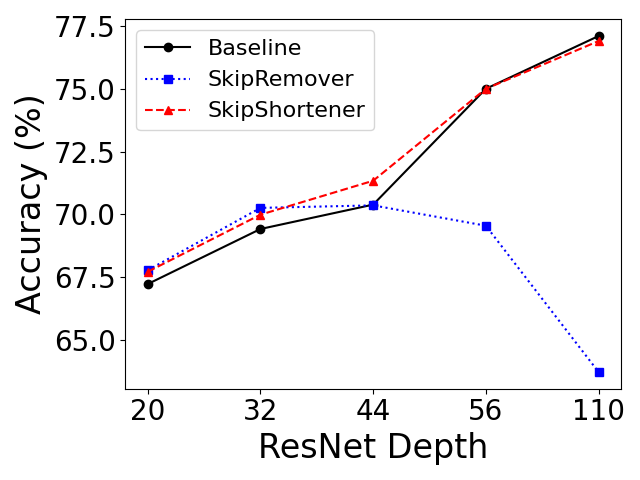}}
    \vspace{1mm}
    \subfloat[SVHN]{\label{fig:svhn}
    \includegraphics[width=0.3\textwidth]{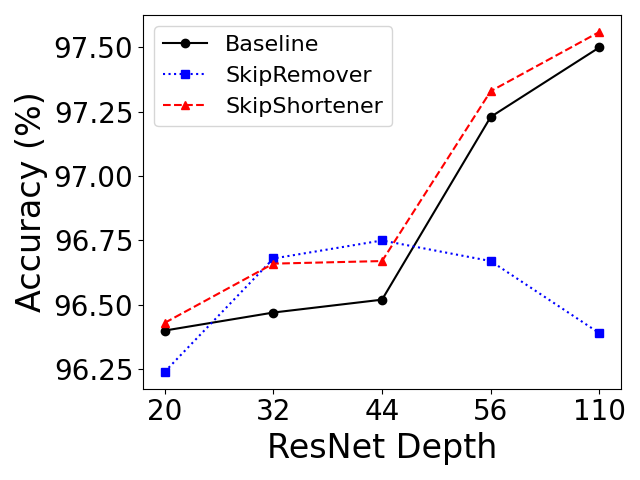}}
    \caption{Top-1 accuracy of \remover and \shortener ResNets of increasing depth on various datasets. ``Baseline'' refers to an unmodified ResNet with conventional skip connections.}
    \label{fig:training}
\end{figure}

\autoref{fig:training} shows that \remover works well for ResNet-44 and smaller, at times even outperforming its baseline (traditional skip connection model). 
However, its accuracy drops as the number of layers increases.
This indicates that shallower NNs do not need skip connections for these classification tasks, but they become more necessary for deeper networks. 
\shortener mostly outperforms the baseline on all three datasets, even on deep models.

\begin{figure}[t]
    \centering
    \subfloat[Skip-less models]{\label{fig:remover-ablation}
    \includegraphics[width=0.35\textwidth]{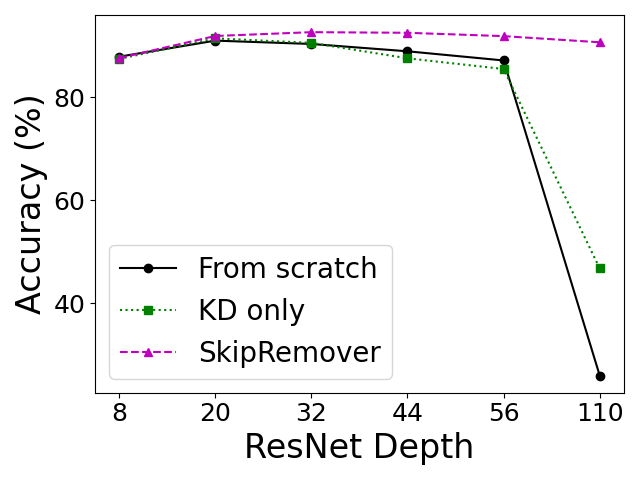}}
    \vspace{1mm}
    \subfloat[Shortened-skip models]{\label{fig:shortener-ablation}
    \includegraphics[width=0.35\textwidth]{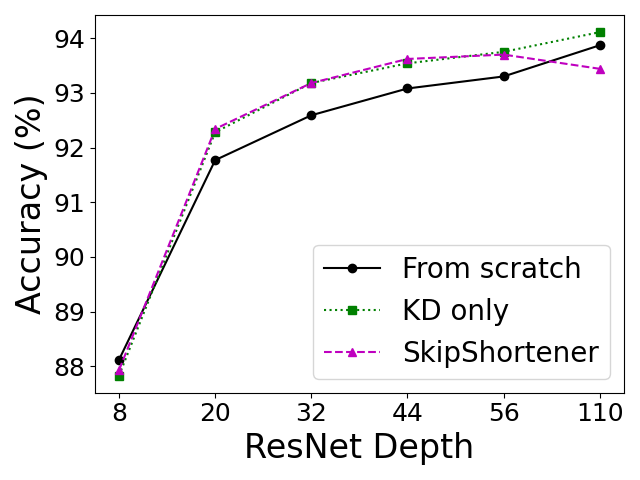}}
    \caption{Accuracy results for ResNets whose skip connections are all altered before training (apart from \remover and \shortener) on CIFAR-10. 
    ``From scratch'' means training with randomly initialized weights without KD. 
    ``KD only'' means training without dynamic skip alterations.} 
    \label{fig:ablation}
\end{figure}

\subsubsection{Ablation Studies}
We also perform \emph{ablation studies} in which we remove key parts of \tool to understand why they are critical to minimizing accuracy loss.
One key part of \remover/\shortener is the dynamic skip connection removal/shortening that occurs every few epochs during training under KD. 
We thus take away this dynamic model alteration by first altering the NNs to have either no skip connections or shortened skip connections.
These pre-modified NNs are then trained under KD only. 
Another key part of \remover and \shortener is KD. 
We evaluate how skip-less and shortened-skip NNs perform without KD, training from randomly initialized weights (i.e., from scratch).

For ResNets trained on CIFAR-10, \remover and \shortener usually yield better results than either normal training or using KD-only on a statically pre-modified network on CIFAR-10 per \autoref{fig:remover-ablation} and \autoref{fig:shortener-ablation}. 
The difference between all of the approaches in the figures is minimal for smaller models, but it becomes more apparent as NN depth increases. 
For instance, skip-less ResNet-110 under regular training yields an accuracy of 26.02\% versus \remover, which achieves an accuracy of 90.68\%, a 64.66\% difference. 
\remover marginally outperforms regular training and KD-only on smaller skip-less models, but performs much better in comparison as the networks deepen. 
\shortener also generally performs better than the other two approaches for shortened skip models. 
Regular training mostly lags behind both KD and \shortener for shortened skip models.

\begin{table}[t]
\caption{\label{tab:resnet-imagenet} Top-1 accuracy of ResNet-50 on the ImageNet dataset. $^\ast$RD~\cite{li2020residual} only removes 82\% of the skip connections.}
\centering
    \begin{tabular}{lc}
        \textbf{Model} & \textbf{Accuracy (\%)} \\
        \hline
        ResNet-50 & 75.85 \\
        \hline
        No skips (from scratch) & 58.36 \\
        No skips (KD only) & 69.40 \\
        Residual distillation (RD)$^\ast$~\cite{li2020residual} & 76.08 \\
        RepVGG-A2~\cite{ding2021repvgg} & 76.48 \\
        \hline
        \textbf{\remover} & \textbf{75.36} \\
    \end{tabular} 
\end{table}

\begin{figure}[t]
    \centering
    \subfloat[ResNet-50 block]{\label{fig:res50-block}
    \includegraphics[width=0.15\textwidth]{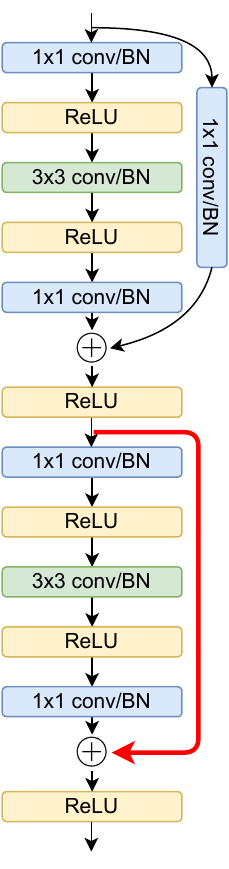}}
    \subfloat[\tool]{\label{fig:tailor-training}
    \includegraphics[width=0.3\textwidth]{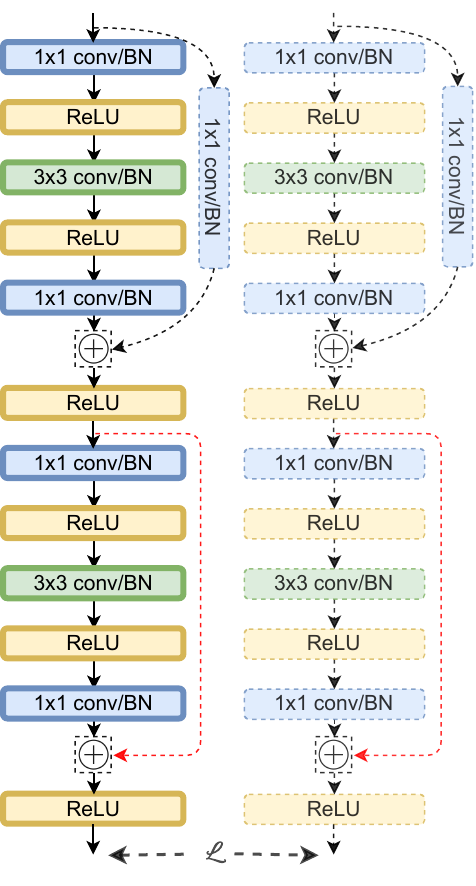}}
    \subfloat[RD~\cite{li2020residual}]{\label{fig:rd-training}
    \includegraphics[width=0.3\textwidth]{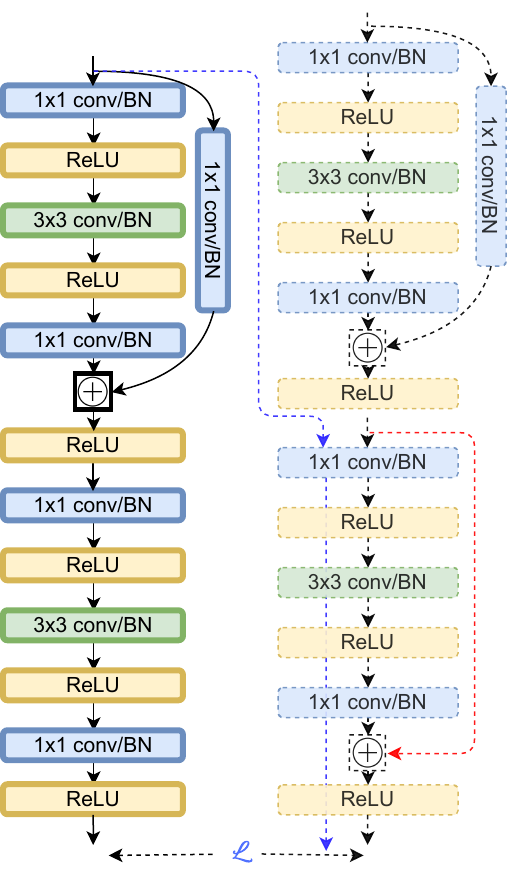}}
    \subfloat[RepVGG~\cite{ding2021repvgg}]{\label{fig:repvgg-training}
    \includegraphics[width=0.25\textwidth]{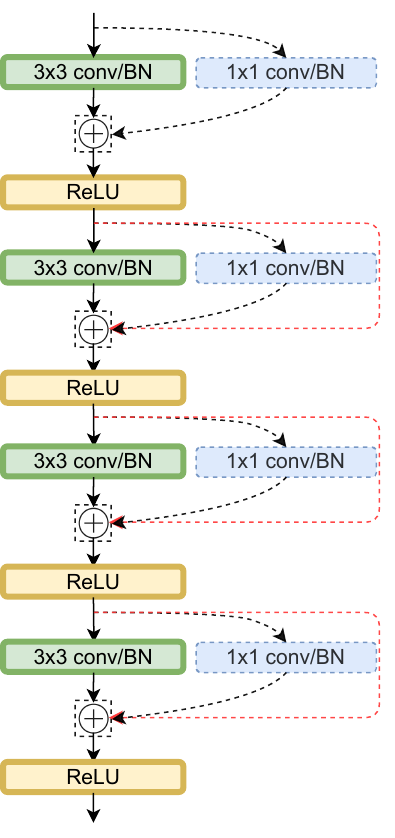}}
    \caption{Comparing \tool's ResNet-50 skip removal method with residual distillation (RD)~\cite{li2020residual} and RepVGG~\cite{ding2021repvgg}. The dashed portions are only used during training and are later removed, leaving the final inference NNs, indicated by bolder lines. Note that \tool (b) removes skip connections from a pretrained ResNet-50 (a). RD does the same but uses a modified KD method that does not remove the 1$\times$1 convolution addition (c). RepVGG starts training from a different NN topology altogether (d).}
    \label{fig:training-comp}
\end{figure}

For ResNet-50 on ImageNet, we only apply \remover because it uses an irregular skip connection architecture known as a ``bottleneck block'' to reduce the number of parameters~\cite{he_deep_2015}. 
This block has a skip connection spanning three layers: a 1$\times$1 convolution, then a 3$\times$3 convolution, then another 1$\times$1 convolution~(\autoref{fig:res50-block}).
This irregular topology is not optimal for \shortener because it requires the majority of the shortened skip connections to pass through extra downsampling 1$\times$1 convolutions to match the activation tensor shapes, significantly increasing the number of model parameters. 
As such, for ResNets with bottleneck blocks, like ResNet-50, we recommend \remover. 
As seen in \autoref{tab:resnet-imagenet}, \remover incurs a 0.49\% accuracy loss compared to the traditional ResNet-50. 
Compared to prior work such as RD~\cite{li2020residual} and RepVGG~\cite{ding2021repvgg}, \remover has slightly lower accuracy (at most 1.12\% accuracy difference)\footnote{Ding et al~\cite{ding2021repvgg} introduce RepVGG models of varying depths. We compare against RepVGG-A2 because it is about the same size as ResNet-50.}.

Nevertheless, \remover has two advantages compared with these methods. 
First, \remover removes all skip connections from ResNet-50, whereas RD only removes 82\% of them.
RD does not remove the 1$\times$1 convolution addition used for downsampling (see \autoref{fig:rd-training}), which is particularly detrimental.
In our experiments on hls4ml architectures, Vivado HLS estimates that ResNet-50's large 1$\times$1 convolution skip connection consumes as many resources as the layers it skips over, effectively doubling resource consumption for that skip connection block.
Although Vivado HLS has a tendency to overestimate the actual place-and-route (P\&R) resource utilization, these estimates demonstrate that performing the 1$\times$1 convolution is a nontrivial task that significantly affects resource consumption.
Second, \remover removes the skip connections from an \textit{existing} pre-trained model, whereas RepVGG requires developers to adopt a new model topology (see \autoref{fig:repvgg-training}).
If developers do not already have a model on hand, RepVGG is a better option. 
However, if developers already have a ResNet trained for their specific dataset, it is advantageous to use \remover if they can afford a small accuracy loss. 
This prevents starting from scratch with RepVGG, which could require extensive hyperparameter tuning.
Even finetuning a pre-trained RepVGG model to a new dataset using transfer learning is time consuming, as it is unclear which of the many methods~\cite{weiss2016survey, zhuang2020comprehensive, pan2009survey} would work best.
Instead, \remover allows developers to take advantage of their existing work and achieve a more resource-efficient model.

\begin{table}[t]
\caption{\label{tab:quartznet-dna} Top-1 accuracy of QuartzNet-5$\times$5 on the Oxford Nanopore Reads dataset.}
\centering
    \begin{tabular}{lc}
        \textbf{Model} & \textbf{Accuracy (\%)} \\
        \hline
        QuartzNet-5$\times$5  & 95.107 \\
        \hline
        No skips (from scratch) & 94.475 \\
        No skips (KD only) & 94.863 \\
        \remover & \textbf{95.086} \\
        Shortened skips (from scratch) & 95.019 \\
        Shortened skips (KD only) & 95.016 \\
        \shortener & 94.902
    \end{tabular} 
\end{table}

\begin{table}[t]
\caption{\label{tab:quartznet-libri} Word error rate (WER) of QuartzNet-10$\times$5 on LibriSpeech dataset. This includes clear (``dev-clean'') and noisy (``dev-other'') audio samples.
``---'' indicates the model failed to converge. } 
\centering
    \begin{tabular}{lcc}
        \multirow{2}{*}{\begin{tabular}[c]{@{}c@{}} \\ \textbf{Model} \end{tabular}} &
        \multirow{2}{*}{\begin{tabular}[c]{@{}c@{}}\textbf{dev-clean} \\ \textbf{WER (\%)}\end{tabular}} & \multirow{2}{*}{\begin{tabular}[c]{@{}c@{}}\textbf{dev-other} \\ \textbf{WER (\%)}\end{tabular}} \\ \\
        \hline
        QuartzNet-10$\times$5 & 5.56 & 16.63 \\
        \hline
        No skips (from scratch) & --- & --- \\
        No skips (KD only) & --- & --- \\
        \remover & --- & --- \\
        Shortened skips (from scratch) & \textbf{6.40} & \textbf{17.68} \\
        Shortened skips (KD only) & 7.14 & 19.95 \\
        \shortener & 7.86 & 21.16
    \end{tabular} 
\end{table}

For QuartzNet-5$\times$5, the \remover model performs the best---only 0.021\% from the baseline (\autoref{tab:quartznet-dna}). 
These results all have high accuracy likely because DNA basecalling is an easier sequence alignment task (only four classes) and the model is more than sufficient.
For a harder ASR task like LibriSpeech, QuartzNet-10$\times$5 fails to converge without skip connections. 
Since the model must translate audio samples to text, the audio samples can be noisy, making ASR harder. 
LibriSpeech, in fact, divides its test samples into ``dev-clean'' for clearly spoken samples and ``dev-other'' for noisy samples.
With such a challenging task, it is not possible to remove the skip connections (like with DNA basecalling).
Nonetheless, QuartzNet-10$\times$5 performs well under \shortener, as it is within 2\% of the baseline WER (\autoref{tab:quartznet-libri}). 
For both QuartzNet-5$\times$5 and -10$\times$5, the best performing shortened skip connection model was one whose skip connections were shortened first and then trained from scratch. 
While \shortener has minimal accuracy loss for both QuartzNets, we recommend training a model with shortened skip connections from scratch for this task.

Overall, \remover and \shortener perform better than either training on randomly initialized weights or training with KD only. 
For harder tasks like ASR though, training a shortened-skip model from scratch is a better choice.
Nevertheless, the success of \remover and \shortener lies in augmenting KD with dynamic skip alterations.

\subsection{Hardware Results}\label{sec:hw-results}
We first quantize ResNets ranging from 20 to 56 layers deep to see how \tool's accuracy fares under reduced precision. 
We then evaluate \tool's effects on hardware resources and latency by performing a case study on ResNet-20-style skip connections implemented using the hls4ml architecture, i.e., the designs illustrated in \autoref{fig:arch}.
We select this style of skip connection because it is the fundamental building block of ResNets that range from 20 to 110 layers.
In our case study, we vary the bit precision and number of filters to see how \tool scales up.
Based on how \tool's resource reductions scale, designers can understand how \tool extrapolates to their own hardware designs.
We report latency as well as P\&R resource results on the Alveo U200 FPGA accelerator card (part no. \texttt{xcu200-fsgd2104-2-e}). 
For end-to-end application results, we evaluate the benefits of \tool on two different styles of CNN architectures.
The first uses the hls4ml tool to generate architectures.
The second is the Reconfigurable DNN Engine---a 2D array of processing elements.
Both styles of architectures are described in \autoref{sec:hw-motivation}.

\subsubsection{Quantization}
\label{sec:quantization} The parameters of a hardware-accelerated NN are typically quantized from floating-point to fixed-point precision~\cite{chang2021mix, yang2019synetgy, moons2017minimum}. 

\begin{figure}[h]
    \centering
    \includegraphics[height=1.5in]{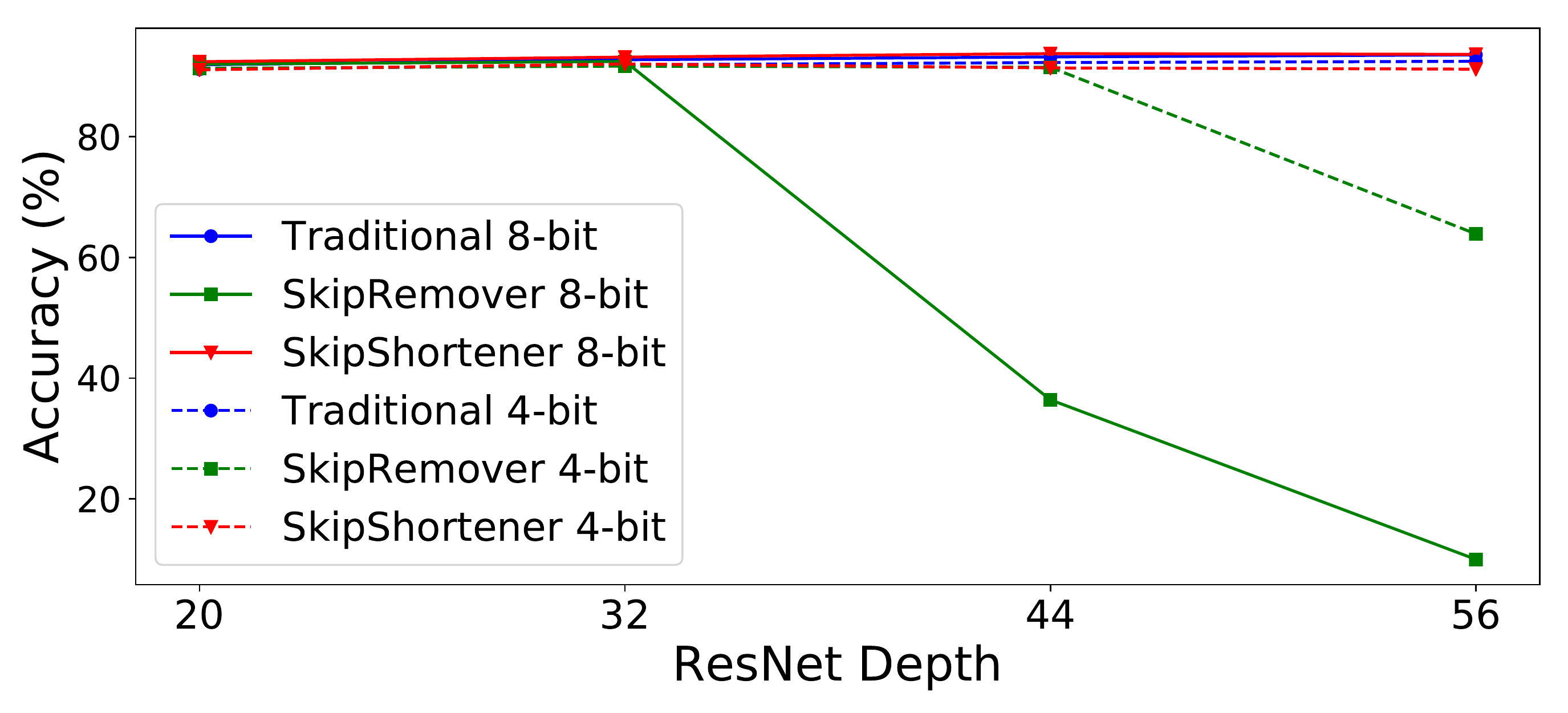}
    \caption{Quantized accuracy results for 8-bit and 4-bit fixed point using Brevitas.}
    \label{fig:quant}
\end{figure}

Quantizing deep NNs with minimal accuracy loss is a largely manual and time-consuming task~\cite{gholami2021survey}. 
We use Brevitas~\cite{brevitas} to quantize our \remover and \shortener ResNets with depths of 20 to 56 from 32-bit floating-point (float32) to 8-bit and 4-bit fixed-point precision on the CIFAR-10 dataset. 
We modified \tool's hardware-aware training algorithm where the teacher continues to use floating-point representation whereas the student is quantized.
This results in the student undergoing quantization-aware training.
In \autoref{fig:quant}, we find that \shortener ResNets consistently outperform traditional ResNets under Brevitas quantization-aware training by 0.5\%. 
\remover ResNets start to suffer from the lack of bits as they get deeper, with accuracy dropping to random classification for ResNet-56. 
But, Brevitas is only one of dozens of ways to quantize neural networks~\cite{moons2017minimum, gholami2021survey, wang2019haq, dong2019hawq, dong2019hawqv2}, so it may be the case that a \remover ResNet-56 requires a different method of quantization to achieve a quantized accuracy similar to its float32 counterpart.

\subsubsection{FPGA Evaluation}\label{sec:fpga_eval}
Our first study looks solely at one ResNet block. 
The second study performs an end-to-end implementation of ResNet8 and ResNet50. 


For our case study on a ResNet skip connection blocks (see designs in \autoref{fig:arch}), we evaluate \tool at \texttt{ap\_fixed<8,3>} and \texttt{ap\_fixed<16,6>} precisions using the hls4ml architecture.
Under both bitwidths, we increase the number of filters for all designs from 16 to 32 to 64.
This way, we can understand how \tool scales with the number of filters.
We use hls4ml~\cite{fahim2021hls4ml} to translate these hardware designs into Vivado HLS, targeting the Alveo U200 FPGA accelerator card. 
hls4ml uses task-level pipelining (i.e., HLS dataflow) for each NN layer, or small group of layers and streams data between dataflow stages using FIFOs. 
hls4ml also exposes a knob known as \textit{reuse factor}, which determines how often multipliers are reused in a design.
To fairly compare our designs as the number of filters increases, we fix the reuse factor to 576.
We then synthesize our designs to report P\&R resource utilization as well as co-simulation latency results.
Lastly, we run the designs on the U200 to verify correctness.

\begin{table*}[t]
\caption{\label{tab:single_block_resource_8bit} Place-and-route resource utilization of a skip connection block as the number of filters increases for $\langle8,3\rangle$ precision on an Alveo U200. 
\remover reduces LUT and FF usage, whereas \shortener trades an increase in FFs for a decrease in LUTs.
T = Traditional, R = \remover, S = \shortener.}
\centering
    \begin{tabular}{c|ccc|ccc|c|c}
        \multirow{2}{*}{\textbf{\# filters}} & & \textbf{LUT} & & & \textbf{FF} & & \textbf{DSP} & \textbf{BRAM} \\
        & T & R & S & T & R & S & T/R/S & T/R/S   \\
        \hline
        16 & 9,984 & 8,482 & 9,764 & 8,654 & 7,841 & 8,916 & 0 & 18.5 \\
        32 & 19,566 & 16,512 & 18,993 & 16,183 & 14,506 & 16,489 & 0 & 36.5 \\
        64 & 42,688 & 36,882 & 42,121 & 31,124 & 27,815 & 31,850 & 0 & 82
    \end{tabular} 
\end{table*}

\begin{figure}[t]
    \centering
    \subfloat[LUT]{\label{fig:filter_scaling_lut_line_8bit}
    \includegraphics[width=0.35\textwidth]{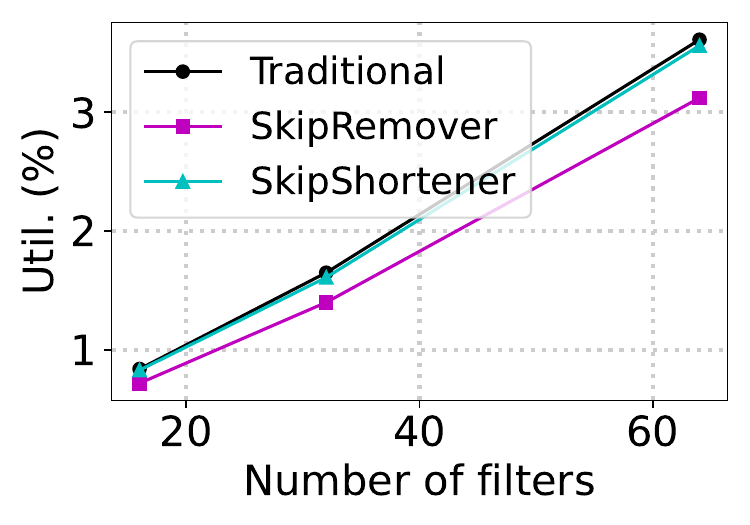}}
    \hspace{1mm}
    \subfloat[FF]{\label{fig:filter_scaling_ff_line_8bit}
    \includegraphics[width=0.35\textwidth]{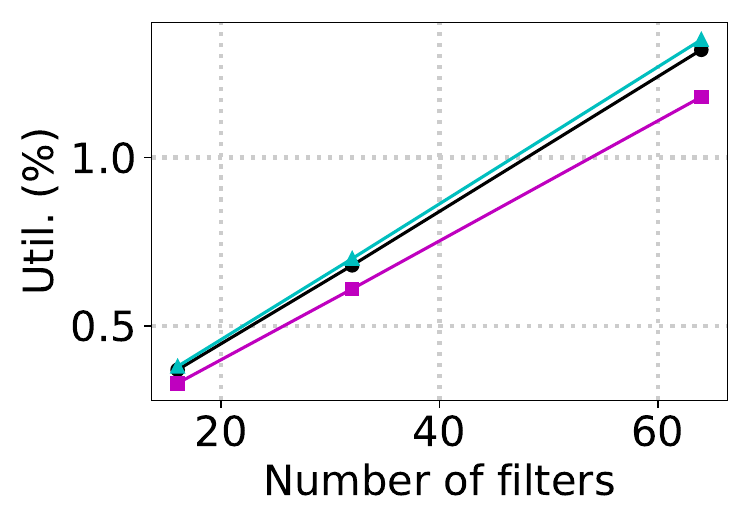}}
    \caption{Percent resource utilization of a $\langle8,3\rangle$ skip connection block at various filter sizes on an Alveo U200. DSPs and BRAMs remain the same across the three designs, so they are not shown. \remover and \shortener LUT and FF reductions scale linearly, as expected.}
    \label{fig:filter_line_8bit}
\end{figure}

\begin{figure}[t]
    \centering
    \subfloat[LUT]{\label{fig:filter_scaling_lut_bar_8bit}
    \includegraphics[width=0.4\textwidth]{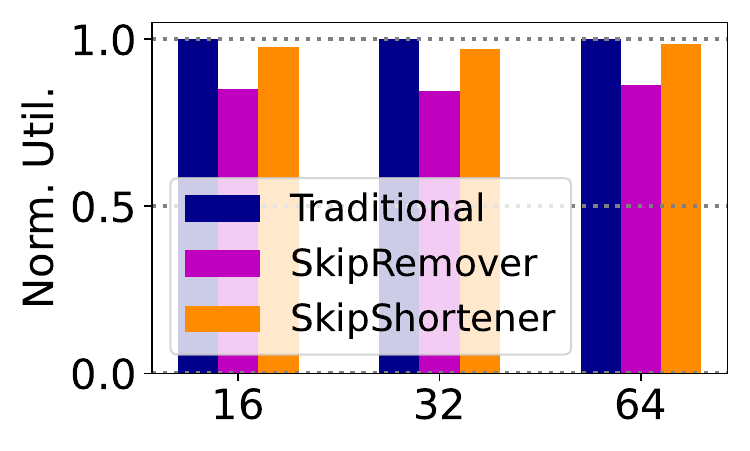}}
    \hspace{1mm}
    \subfloat[FF]{\label{fig:filter_scaling_ff_bar_8bit}
    \includegraphics[width=0.4\textwidth]{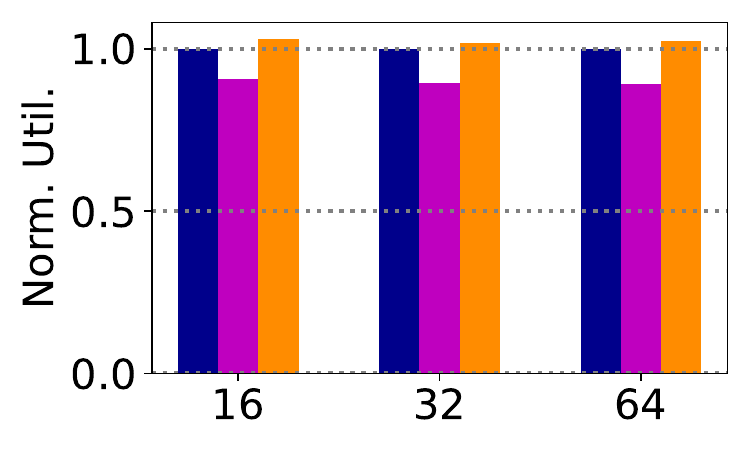}}
    \caption{Resource utilization normalized to the traditional design of a $\langle8,3\rangle$ skip connection block at various filter sizes. DSPs and BRAMs remain the same across the three designs, so they are not shown. \remover and \shortener LUT and FF reductions scale proportionally, as expected.}
    \label{fig:filter_bar_8bit}
\end{figure}

Under 8-bit precision, we find that both \remover and \shortener reduce resources. 
\autoref{tab:single_block_resource_8bit} summarizes our P\&R results.
Since our model uses 8-bit precision, we see that all of our models exhibit low DSP usage and higher LUT and FF utilization. 
This is because Vivado HLS maps multiplications on datatypes that are less than 10 bits to LUTs instead of DSPs, as noted by~\cite{aarrestad2021fast, yang2019synetgy}. 
It is possible to pack two 8-bit weights into a DSP~\cite{fu2016deep}, but this is out of scope and orthogonal to the effects \tool has on hardware. 
Furthermore, all of the traditional and \tool designs use the same amount of BRAMs with respect to the number of filters because here the BRAMs are used solely for on-chip weight storage, which does not differ across design. 
Nonetheless, \remover decreases LUT usage by up to 16\% and FF usage by up to 11\% compared with the traditional design (\autoref{fig:filter_bar_8bit}).
These resource savings represent the extra hardware needed to implement a skip connection and subsequently the resources saved.
As previously mentioned in \autoref{sec:hw-motivation}, the extra dataflow stages that carry out a skip connection are no longer necessary.
More importantly, \remover's savings scale linearly as the number of filters increases from 16 to 64 (\autoref{fig:filter_line_8bit}). 
\shortener's resource reductions present a tradeoff, increasing FFs by 2\% in exchange for decreasing LUTs by 3\% (\autoref{fig:filter_bar_8bit}).
\shortener lowers LUT utilization because the lifespan of each skip connection lasts only one dataflow stage instead of the traditional two. 
This means we need not spend extra logic on the dataflow stages needed to copy the skip connections to buffers that last longer than one stage.
However, since the shortened skip connection now fully resides in a single dataflow stage (previously described in \autoref{fig:short-arch}), this requires some extra FFs.
This represents the tradeoff \shortener provides at 8-bit precision: some extra FFs for fewer LUTs.
These resource tradeoffs also scale linearly as the number of filters scales up, as seen in \autoref{fig:filter_line_8bit}.
 
We find more dramatic resource reductions when we look at our 16-bit designs.
\autoref{tab:single_block_resource_16bit} summarizes our P\&R results.
In contrast with our 8-bit designs, at higher precision, our designs rely more on DSPs and BRAMs. 
This time the BRAMs are used not only to store weights on chip but also to implement the FIFOs that connect the dataflow stages.
Therefore, as we tailor the dataflow stages according to each design (e.g., \remover or \shortener), the BRAMs now also reflect these changes.
At its best, \remover lowers LUTs by 11\%, FFs by 13\%, and BRAMs by 13\%.
Without a skip connection to implement, \remover uses fewer resources than the traditional design.
The DSPs remains unchanged because they are used solely for the convolutional layers' multiplications and not the skip connection, which is also the case for \shortener.

\begin{figure}[t]
    \centering
    \subfloat[LUT]{\label{fig:filter_scaling_lut_16bit}
    \includegraphics[width=0.3\textwidth]{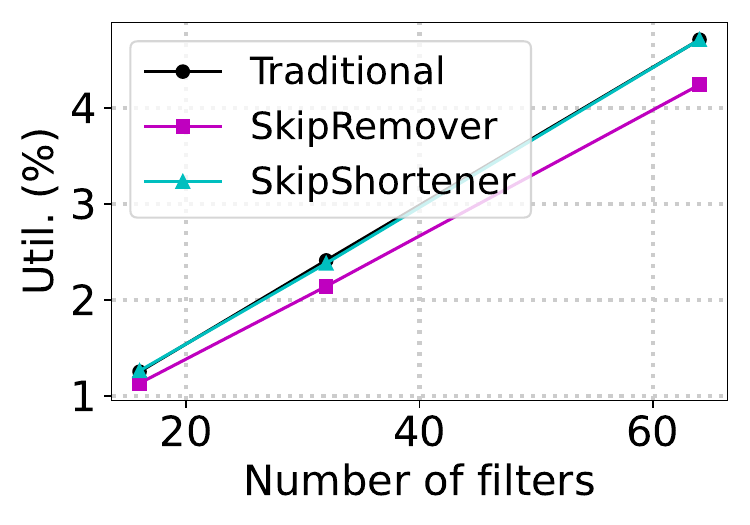}}
    \vspace{1mm}
    \subfloat[FF]{\label{fig:filter_scaling_ff_16bit}
    \includegraphics[width=0.3\textwidth]{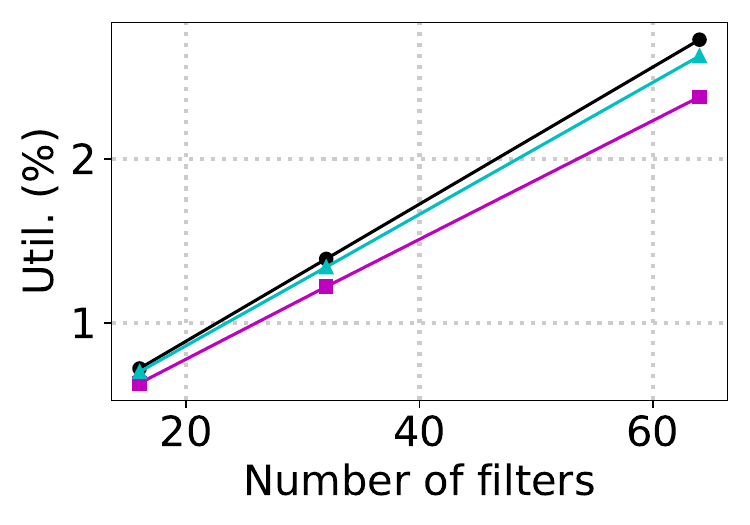}}
    \hspace{1mm}
    \subfloat[BRAM]{\label{fig:filter_scaling_bram_16bit}
    \includegraphics[width=0.3\textwidth]{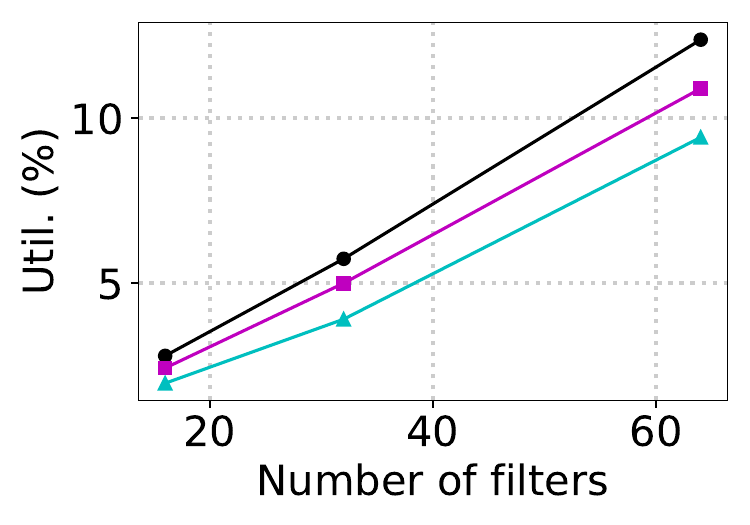}}
    \caption{Percent resource utilization of a $\langle16,6\rangle$ skip connection block at various filter sizes on an Alveo U200. 
    \remover and \shortener resource reductions scale linearly, as expected.}
    \label{fig:filter_line_16bit}
\end{figure}

Similar to the 8-bit designs, \shortener presents a resource tradeoff---this time trading a small increase in LUTs (at most 1\%) for decreases in FFs and BRAMs.
In the best case, \shortener reduces LUTs by 1\%, FFs by 4\%, and BRAMs by 34\%. 
While \shortener uses fewer LUTs than the traditional case for 32 filters, \shortener pays about a 1\% increase in LUTs for 16 and 64 filters in exchange for decreases in FFs and BRAMs.
This small disparity is likely an artifact of the heuristics Vivado P\&R uses to allocate resources.
Again, these resource tradeoffs and savings are possible because the shortened skip connections can be implemented within a single dataflow stage due to its reduced lifetime.
\autoref{tab:fifo_depth} shows that the lifetime of each shortened skip connection is a little less than half the lifetime of the traditional one.
With shorter lifetimes, we find that the \shortener's skip connections' FIFOs can now be implemented using shift registers instead of BRAMs, which is what the traditional design still uses (\autoref{tab:fifo_depth}).
Shift registers are much more efficient memories compared to BRAMs.
As such, it is advantageous to hardware designers to consider how \shortener provides opportunity to implement skip connections with a more efficient memory architecture like shift registers.
This leads to 30--34\% fewer BRAMs than the traditional design, even as the number of filters scales up.
While in this case \shortener uses fewer BRAMs than \remover does, \shortener offsets this difference by using more FFs than \remover does. 
For both \remover and \shortener, resource utilization (and the associated reductions) scale linearly, as seen in \autoref{fig:filter_line_16bit}.

\begin{table*}[t]
\caption{\label{tab:single_block_resource_16bit} Place-and-route resource utilization of a skip connection block as the number of filters increases for $\langle16, 6\rangle$ precision on an Alveo U200. 
\remover reduces resources across the board, whereas \shortener trades an increase in LUTs for a decrease in FFs and BRAMs.
T = Traditional, R = \remover, S = \shortener.}
\centering
    \begin{tabular}{c|ccc|ccc|c|ccc}
        \multirow{2}{*}{\textbf{\# filters}} & & \textbf{LUT} & & & \textbf{FF} & & \textbf{DSP} & & \textbf{BRAM} \\
        & T & R & S & T & R & S & T/R/S & T & R & S   \\
        \hline
        16 & 14,733 & 13,320 & 14,933 & 17,044 & 14,935 & 16,438 & 12 & 60.5 & 52.5 & 42.5 \\
        32 & 28,498 & 25,330 & 28,184 & 32,923 & 28,747 & 31,764 & 48 & 124 & 108 & 84.5 \\
        64 & 55,699 & 50,074 & 55,720 & 64,564 & 56,263 & 62,252 & 192 & 267.5 & 235.5 & 203.5 
    \end{tabular} 
\end{table*}

\begin{figure}[t]
    \centering
    \includegraphics[width=0.9\textwidth]{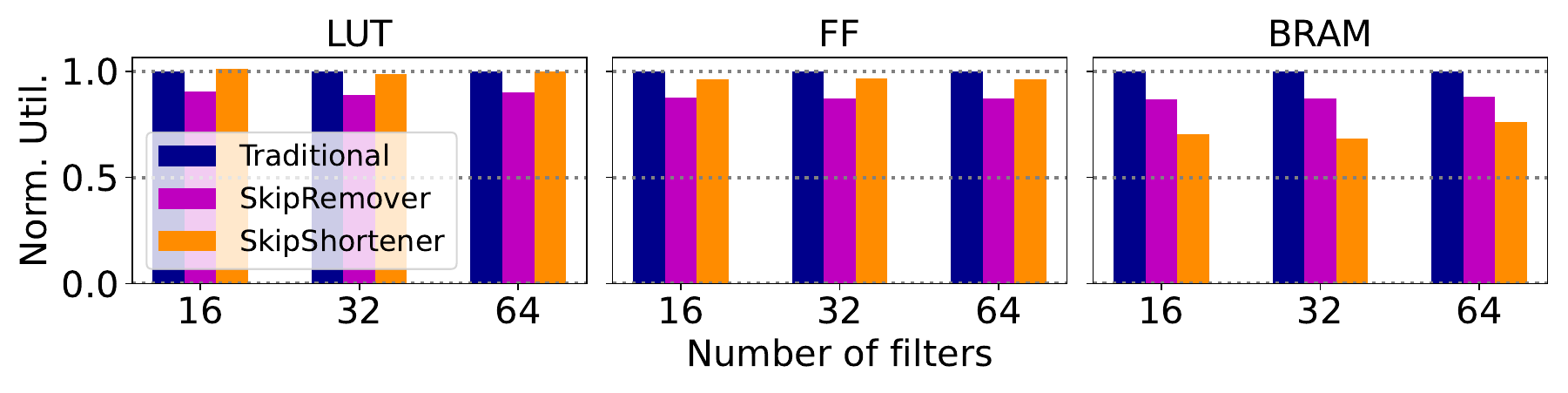}
    \caption{Resource utilization normalized to the traditional design of a $\langle16,6\rangle$ skip connection block at various filter sizes. The \remover and \shortener resource savings scale proportionally as the number of filters scales up.}
    \label{fig:filter_bar_16bit}
\end{figure}

\begin{table}[t]
\caption{\label{tab:fifo_depth} FIFO depths of a single skip connection hardware design at 16-bit precision. 
\remover has no skip connections, so it has no skip connection FIFOs.}
\centering
    \begin{tabular}{l|cc}
        Hardware Design & FIFO Depth & FIFO Implementation \\
        \hline
        Traditional & 69 & BRAM \\
        \remover & 0 & --- \\
        \shortener 1st skip & 33 & Shift Register \\
        \shortener 2nd skip & 34 & Shift Register \\
    \end{tabular} 
\end{table}

\begin{table}[t]
\caption{Latency co-simulation results of a skip connection block at $\langle8,3\rangle$ and $\langle16, 6\rangle$ precision. 
The latency for the Traditional, \remover, and \shortener designs are the same for each number of filters because they all rely on task-level pipelining that reuses multipliers at the same rate (576$\times$).}
    \label{tab:latency}
    \centering
    \begin{tabular}{c|c}
        \multirow{2}{*}{\textbf{\# filters}}  & Latency (ms)  \\
        & Traditional/\remover/\shortener \\
        \hline
         16 & 23.38 \\ 
         32 & 23.05 \\
         64 & 22.39
    \end{tabular}
\end{table}


\tool does not affect latency for hls4ml architectures. 
As seen in \autoref{tab:latency}, for each number of filters, all designs exhibit the same latency, according to co-simulation on an Alveo U200. 
The slight decrease in latency as the number of filters scales is due to an increase in DSPs and a higher degree of parallelism.
As discussed in \autoref{sec:hw-motivation}, hls4ml designs pipeline their tasks.
The convolutions' multiplication tasks dominate the overall dataflow latency.
The tasks that \remover eliminates and \shortener implements more efficiently, namely the skip connection cloning and addition stages, have significantly lower latency than the convolutions and are thus not on the critical path. 
The throughput thus remains the same.

By shortening skip connections, we reduce their lifespans, which provides an opportunity for simplifying their hardware implementation specifically for hls4ml architectures. 
However, shortening skip connections is not beneficial for all architectures.
As seen in \autoref{tab:throughput}, shortening skip connections is worse for both GPU and CPU because doing so increases off-chip memory accesses.
These extra accesses lower throughput by 5\% on GPU and 2\% on CPU. 
On FPGAs with hls4ml architectures, however, we can modify the architecture to take advantage of shortened skip connections, reducing resource consumption without negatively affecting throughput (\autoref{tab:throughput}).

\begin{table}[h]
\caption{\label{tab:throughput} Normalized throughput of a ResNet20. The GPU and CPU both were run with batch size = 64, whereas FPGA was run with batch size = 1. Throughput is normalized column-wise to the top entry. GPU = 1080Ti. CPU = AMD Ryzen 9 5900X. FPGA = Alveo U200. \remover increases GPU and CPU throughput because it decreases off-chip memory accesses. \shortener, however, decreases GPU and CPU throughput because it increases off-chip memory accesses. For a fully on-chip, dataflowed FPGA architecture, neither \remover nor \shortener have any effect on throughput.}
\centering
    \begin{tabular}{l|r|r|r}
        \textbf{Model} & \textbf{GPU} & \textbf{CPU} & \textbf{FPGA} \\
        \hline
        Traditional skip connections & 1$\times$ & 1$\times$ & 1$\times$ \\
        \remover & 1.11$\times$ & 1.03$\times$ & 1$\times$ \\
        \shortener & 0.95$\times$ & 0.98$\times$ & 1$\times$ \\
    \end{tabular} 
\end{table}

We performed two studies to understand how \tool performs for end-to-end implementations of ResNet models.
The first is ResNet8 from  MLPerf Tiny that was designed in hls4ml~\cite{borras2022open, banbury2021mlperf}. 
The second is ResNet50 implemented on the Reconfigurable DNN architecture. 

The ResNet8 model targets the Alveo U200.
It uses 16-bit fixed-point representation with six integer bits.
The reuse factor for the layers was hand-tuned to 72, which directly affects the resource usage and latency of the layers.
The reuse factor is one of the more important knobs for design space exploration in hls4ml and is often hand-tuned to maximize resource usage of the platform while optimizing the overall network performance. 

\begin{table}[h]
\caption{MLPerf Tiny ResNet8 model implemented using hls4ml with skip connection, with shortened skip connections, and without skip connections. }
    \label{tab:resnet8-hls4ml}
\centering
\begin{tabular}{l|c|c|c}
 & \textbf{With Skip Connections}
 & \textbf{Shortened Skip Connections} 
 & \textbf{Without Skip Connections} \\
        \hline
        Accuracy (\%)       & 87.39                    & 87.93  & 87.62\\
        LUTs           & 158609                        & 165699 &  144206\\
        FFs                  & 196012                         & 204914 & 181768 \\
        DSP48s & 1083                         & 1083 & 1043\\
        BRAMs & 173 & 158.5 & 156
    \end{tabular} 
\end{table}

\autoref{tab:resnet8-hls4ml} shows the resource usage results for the ResNet8 model with skip connections, with shortened skip connections, and without skip connections. 
Removing the skip connections has clear benefits across all the resources.
Shortening the skip connections reduces BRAMs while increasing LUTs and FFs. 
Both the shortened skip connections and the removed skip connections models show improved accuracy over traditional skip connections.
In all cases, the latency remains the same, requiring 304,697 cycles running at 100 MHz (approximately 3ms/inference).

Our second full model case study implemented a Reconfigurable DNN architecture on the ZCU102 development board which contains a Zynq UltraScale+ MPSoC. 
The Reconfigurable DNN array is configured to have 7 rows $\times$ 96 columns for a total of 672 of processing elements (PEs) that support 8-bit inputs and 8-bit weights. 
Each PE contains a multiplier and an accumulator implemented using DSPs on FPGA fabric. 
Input pixels and weights are streamed into the engine as AXI-Stream packets. 
Images are processed in batches of 7, to increase the reuse and reduce memory accesses.
The Reconfigurable DNN architecture was synthesized, placed \& routed at a clock frequency of 250 MHz on a ZCU102. The architecture with $7 \times 96 = 672$ PEs used 49057 LUTs (18\%), 81446 flip flops (15\%), 114 BRAMs (13\%), and 1344 DSPs (53\%) on the FPGA fabric.

We implemented a ResNet50 model with and without skip connections on a 672-element Reconfigurable DNN architecture running on the ZCU102. 
\autoref{tab:dnn_engine_perfomance} shows the performance of ResNet50.
Removing the skip connections largely benefits the performance due to the removal of the $1 \times 1$ convolution blocks. 
Removing the skip connections also removes those layers, which no longer need to be scheduled on the PE array. 
The results are much better performance in terms of all metrics: approximately 30\% increases in FPS and latency and approximately 45\% decrease in memory accesses.

\begin{table*}[h]
\caption{\label{tab:dnn_engine_perfomance}  ResNet50 performance with and without skip connections on the Reconfigurable DNN architecture. The architecture has  672 processing elements and runs on the ZCU102 development board at 250 MHz.}
\centering
    \begin{tabular}{l|c|c}
                                     & \textbf{With skip connections} &\textbf{Without skip connections} \\
        \hline
        Accuracy (\%)       & 75.85                    & 75.36  \\
        Frames per second (FPS)           & 28.69                          & 37.47 \\
        Time per image (s)           & 0.035                          & 0.027 \\
        Latency (s)                  & 0.244                          & 0.187 \\
        Memory access per image (Mb) & 140.95                         & 92.71 
    \end{tabular} 
\end{table*}

\section{Discussion}
\label{sec:discussion}

With these results in hand, designers can now consider which accuracy versus resource tradeoffs they are willing to make during the hardware-software codesign process. 

\remover provides minimal accuracy loss while reducing resource consumption and increasing performance---a win-win scenario. 
As seen in \autoref{sec:training}, \remover ResNet-50 is only 0.49\% less accurate than the baseline on ImageNet. 
But, \remover is less effective on deeper NNs (such as QuartzNet-10$\times$5 and ResNet-110). 
In fact, QuartzNet-10$\times$5 fails to converge when trained under \remover. 
For such deep NNs trained on difficult tasks like ASR, skip connections are instrumental in training convergence~\cite{he_deep_2015}.
By removing skip connections, we expect and see a degradation in accuracy for deeper NNs.
This degradation is not as drastic for other tasks.
For instance, ResNet-110 still converges when trained using \remover, but it is 3.72\% less accurate on CIFAR-10 and 9.61\% less accurate on CIFAR-100, compared to the original baseline model.
We propose this tradeoff between NN size and \remover performance as an additional consideration during design space exploration. 
In response, \shortener is more suitable for deeper NNs when \remover is less effective. 
\shortener maintains accuracy comparable to its original skip connection models and reduces resource requirements by up to 34\% compared to the traditional skip connection model. 

Based on our hls4ml evaluation, designers can extrapolate to their own designs because, as we have shown in \autoref{fig:filter_line_8bit} and \autoref{fig:filter_line_16bit}, 
the resource usage and savings scale linearly as the number of filters grows.
We have also shown that at the higher 16-bit precision, \tool provides significant resource reductions, so if designers need more precision, \tool's savings will follow.
If they need lower 8-bit precision, \remover still manages to lower the 8-bit designs' LUTs by 16\% and FFs by 11\%.
Even \shortener decreases LUTs by 3\% despite a 2\% increase in FFs, though these smaller resource savings are offset by its overall higher accuracy performance compared with \remover.
As a result, it is up to the designer to consider how to best apply \tool's codesign methods given their accuracy and resource requirements.

\subsection{Theoretical Understanding}
Prior work investigated why skip connections are so helpful to ResNets. 
Veit et al~\cite{veit2016residual} argue that ResNets behave like ensembles of smaller subnetworks that vary in depth and allow the NN to train and converge more easily. 
Li et al~\cite{li2018visualizing} and Yao et al~\cite{yao2020pyhessian} show that introducing skip connections make the NN loss landscape to be much smoother and have less nonconvexity.
They show that naively removing these skip connections causes an explosion of nonconvexity in the loss landscape, which makes training significantly more difficult. 
We confirm these results in our ablation studies~(\autoref{sec:training}), as accuracy indeed drops when skip connections are removed naively. 
With both KD and SkipRemover, we see an improvement in accuracy.
Since the student is trying to mimic the teacher's outputs, it is possible that the teacher's outputs guide the student in such a way that prevents the loss landscape from becoming less smooth.
Theoretical work from Lin et al~\cite{lin2018resnet} has proven that a ResNet with one-neuron hidden layers is a universal approximator.
This work suggests that adding more neurons to the hidden layers creates an over-parameterized ResNet.
Since stochastic gradient descent performs better in the presence of over-parameterization, having more neurons per hidden layer increases training efficiency, making it easier to converge.
This work also argues that a ResNet is essentially a sparse version of a fully connected NN because the identity skip connections create simpler paths within the ResNet, which was similarly posited by Lin et al~\cite{lin2018resnet}.
Given that CNNs and ResNets have both been proven to be universal approximators~\cite{petersen2020equivalence, lin2018resnet}, this implies that there exists a set of parameters for a CNN that can mimic a ResNet such that they equal the same function.
It is mainly easier to find a well performing ResNet because Lin et al~\cite{lin2018resnet} showed that one-neuron hidden layers is sufficient for a ResNet to be a universal approximator. 

\subsection{Future Work}
In our work, \tool has taken removing and shortening skip connections to their extremes: it either fully removes or fully shortens all the skip connections in a NN.
It would be worthwhile to understand the accuracy versus resource utilization tradeoff under less extreme cases, e.g., removing only half of the skip connections. 
It would also be interesting to mix \remover and \shortener together to try and recover accuracy in the instances when \remover fails.
These approaches may help address \remover's scalability issues and strike a balance between \shortener's high accuracy and \remover's resource savings and performance improvements.

\section{Conclusion}
\label{sec:conclusion}
\tool introduces two new methods, \remover and \shortener, that alters NNs with skip connections dynamically during retraining to fit better on hardware, achieving resource-efficient inference with minimal to no loss in accuracy.
With \remover, NNs no longer need to rely on skip connections for high accuracy during inference. 
With \shortener, we retrain NNs to use shorter skip connections with minimal to no loss in accuracy. 
Shortening skip connections is beneficial for hardware architectures generated by the hls4ml tool as it reduces the skip connection lifetime. 
We demonstrate FPGA resource consumption reductions of up to 34\% for BRAMs, 13\% for FFs, and 16\% for LUTs.
We show that \tool is also valuable for optimizing 2D PE array architectures. \remover increases performance by 30\% and decreases memory bandwidth by 45\%. 
Designers can decide which accuracy versus resource tradeoffs offered by \remover and \shortener are suitable to their design requirements. 
As a result, \tool is another tool in the hardware-software codesign toolbox for designers to use when building customized accelerators. 


 \begin{acks}
    The authors thank the anonymous referees for their valuable comments and helpful suggestions.
    This material is based upon work supported by the National Science Foundation Graduate Research Fellowship Program under Grant No. DGE-2038238. 
    Any opinions, findings, and conclusions or recommendations expressed in this material are those of the author(s) and do not necessarily reflect the views of the National Science Foundation.
    This work was also partially supported by the \grantsponsor{DOE}{U.S. Department of Energy (DOE)}{https://doi.org/10.13039/100000015}, Office of Science, Office of Advanced Scientific Computing Research under the ``Real-time Data Reduction Codesign at the Extreme Edge for Science'' Project (\grantnum{DOE}{DE-FOA-0002501}).
    JD was also supported by the DOE Office of Science, Office of High Energy Physics Early Career Research Program under award number \grantnum{DOE}{DE-SC0021187} and the \grantsponsor{NSF}{National Science Foundation (NSF)}{https://doi.org/10.13039/100000001} under award number \grantnum[https://a3d3.ai]{NSF}{2117997}.
\end{acks}
\bibliographystyle{ACM-Reference-Format}
\bibliography{refs}


\begin{thebibliography}{52}


\ifx \showCODEN    \undefined \def \showCODEN     #1{\unskip}     \fi
\ifx \showDOI      \undefined \def \showDOI       #1{#1}\fi
\ifx \showISBNx    \undefined \def \showISBNx     #1{\unskip}     \fi
\ifx \showISBNxiii \undefined \def \showISBNxiii  #1{\unskip}     \fi
\ifx \showISSN     \undefined \def \showISSN      #1{\unskip}     \fi
\ifx \showLCCN     \undefined \def \showLCCN      #1{\unskip}     \fi
\ifx \shownote     \undefined \def \shownote      #1{#1}          \fi
\ifx \showarticletitle \undefined \def \showarticletitle #1{#1}   \fi
\ifx \showURL      \undefined \def \showURL       {\relax}        \fi
\providecommand\bibfield[2]{#2}
\providecommand\bibinfo[2]{#2}
\providecommand\natexlab[1]{#1}
\providecommand\showeprint[2][]{arXiv:#2}

\bibitem[Tai(2023)]%
        {Tailor}
 \bibinfo{year}{2023}\natexlab{}.
\newblock \bibinfo{title}{Tailor}.
\newblock \bibinfo{howpublished}{\url{https://github.com/oliviaweng/tailor}}.
\newblock


\bibitem[Aarrestad et~al\mbox{.}(2021)]%
        {aarrestad2021fast}
\bibfield{author}{\bibinfo{person}{Thea Aarrestad} {et~al\mbox{.}}}
  \bibinfo{year}{2021}\natexlab{}.
\newblock \showarticletitle{{Fast convolutional neural networks on FPGAs with
  hls4ml}}.
\newblock \bibinfo{journal}{\emph{Mach. Learn.: Sci. Technol.}}
  \bibinfo{volume}{2}, \bibinfo{number}{4} (\bibinfo{year}{2021}),
  \bibinfo{pages}{045015}.
\newblock
\urldef\tempurl%
\url{https://doi.org/10.1088/2632-2153/ac0ea1}
\showDOI{\tempurl}
\showeprint[arxiv]{2101.05108}~[cs.LG]


\bibitem[Banbury et~al\mbox{.}(2021)]%
        {banbury2021mlperf}
\bibfield{author}{\bibinfo{person}{Colby Banbury},
  \bibinfo{person}{Vijay~Janapa Reddi}, \bibinfo{person}{Peter Torelli},
  \bibinfo{person}{Jeremy Holleman}, \bibinfo{person}{Nat Jeffries},
  \bibinfo{person}{Csaba Kiraly}, \bibinfo{person}{Pietro Montino},
  \bibinfo{person}{David Kanter}, \bibinfo{person}{Sebastian Ahmed},
  \bibinfo{person}{Danilo Pau}, {et~al\mbox{.}}}
  \bibinfo{year}{2021}\natexlab{}.
\newblock \showarticletitle{Mlperf tiny benchmark}.
\newblock \bibinfo{journal}{\emph{arXiv preprint arXiv:2106.07597}}
  (\bibinfo{year}{2021}).
\newblock


\bibitem[Bengio et~al\mbox{.}(1994)]%
        {bengio_learning_1994}
\bibfield{author}{\bibinfo{person}{Y. Bengio}, \bibinfo{person}{P. Simard},
  {and} \bibinfo{person}{P. Frasconi}.} \bibinfo{year}{1994}\natexlab{}.
\newblock \showarticletitle{Learning long-term dependencies with gradient
  descent is difficult}.
\newblock \bibinfo{journal}{\emph{IEEE Trans. Neural Netw.}}
  \bibinfo{volume}{5}, \bibinfo{number}{2} (\bibinfo{date}{March}
  \bibinfo{year}{1994}), \bibinfo{pages}{157}.
\newblock
\showISSN{1045-9227, 1941-0093}
\urldef\tempurl%
\url{https://doi.org/10.1109/72.279181}
\showDOI{\tempurl}


\bibitem[Borras et~al\mbox{.}(2022)]%
        {borras2022open}
\bibfield{author}{\bibinfo{person}{Hendrik Borras} {et~al\mbox{.}}}
  \bibinfo{year}{2022}\natexlab{}.
\newblock \showarticletitle{Open-source FPGA-ML codesign for the MLPerf Tiny
  Benchmark}. In \bibinfo{booktitle}{\emph{Workshop on Benchmarking Machine
  Learning Workloads on Emerging Hardware (MLBench)}}.
\newblock
\showeprint{2206.11791}~[cs.LG]


\bibitem[Chang et~al\mbox{.}(2021)]%
        {chang2021mix}
\bibfield{author}{\bibinfo{person}{Sung-En Chang}, \bibinfo{person}{Yanyu Li},
  \bibinfo{person}{Mengshu Sun}, \bibinfo{person}{Runbin Shi},
  \bibinfo{person}{Hayden K-H So}, \bibinfo{person}{Xuehai Qian},
  \bibinfo{person}{Yanzhi Wang}, {and} \bibinfo{person}{Xue Lin}.}
  \bibinfo{year}{2021}\natexlab{}.
\newblock \showarticletitle{Mix and Match: A novel {FPGA}-centric deep neural
  network quantization framework}. In \bibinfo{booktitle}{\emph{2021 IEEE
  International Symposium on High-Performance Computer Architecture (HPCA)}}.
  IEEE, \bibinfo{pages}{208}.
\newblock
\urldef\tempurl%
\url{https://doi.org/10.1109/HPCA51647.2021.00027}
\showDOI{\tempurl}
\showeprint{2012.04240}


\bibitem[Deng et~al\mbox{.}(2009)]%
        {deng2009imagenet}
\bibfield{author}{\bibinfo{person}{Jia Deng}, \bibinfo{person}{Wei Dong},
  \bibinfo{person}{Richard Socher}, \bibinfo{person}{Li-Jia Li},
  \bibinfo{person}{Kai Li}, {and} \bibinfo{person}{Li Fei-Fei}.}
  \bibinfo{year}{2009}\natexlab{}.
\newblock \showarticletitle{{ImageNet}: A large-scale hierarchical image
  database}. In \bibinfo{booktitle}{\emph{2009 IEEE conference on computer
  vision and pattern recognition}}. \bibinfo{pages}{248}.
\newblock
\urldef\tempurl%
\url{https://doi.org/10.1109/CVPR.2009.5206848}
\showDOI{\tempurl}


\bibitem[Ding et~al\mbox{.}(2021)]%
        {ding2021repvgg}
\bibfield{author}{\bibinfo{person}{Xiaohan Ding}, \bibinfo{person}{Xiangyu
  Zhang}, \bibinfo{person}{Ningning Ma}, \bibinfo{person}{Jungong Han},
  \bibinfo{person}{Guiguang Ding}, {and} \bibinfo{person}{Jian Sun}.}
  \bibinfo{year}{2021}\natexlab{}.
\newblock \showarticletitle{{RepVGG}: Making {VGG}-style convnets great again}.
  In \bibinfo{booktitle}{\emph{Proc. IEEE/CVF Conference on Computer Vision and
  Pattern Recognition}}. \bibinfo{pages}{13733}.
\newblock
\urldef\tempurl%
\url{https://doi.org/10.1109/CVPR46437.2021.01352}
\showDOI{\tempurl}
\showeprint{2101.03697}


\bibitem[Dong et~al\mbox{.}(2020)]%
        {dong2019hawqv2}
\bibfield{author}{\bibinfo{person}{Zhen Dong}, \bibinfo{person}{Zhewei Yao},
  \bibinfo{person}{Daiyaan Arfeen}, \bibinfo{person}{Amir Gholami},
  \bibinfo{person}{Michael~W Mahoney}, {and} \bibinfo{person}{Kurt Keutzer}.}
  \bibinfo{year}{2020}\natexlab{}.
\newblock \showarticletitle{{HAWQ-V2}: Hessian Aware trace-Weighted
  Quantization of Neural Networks}.
\newblock   \bibinfo{volume}{33} (\bibinfo{year}{2020}),
  \bibinfo{pages}{18518}.
\newblock
\showeprint{1911.03852}
\urldef\tempurl%
\url{https://proceedings.neurips.cc/paper/2020/file/d77c703536718b95308130ff2e5cf9ee-Paper.pdf}
\showURL{%
\tempurl}


\bibitem[Dong et~al\mbox{.}(2019)]%
        {dong2019hawq}
\bibfield{author}{\bibinfo{person}{Zhen Dong}, \bibinfo{person}{Zhewei Yao},
  \bibinfo{person}{Amir Gholami}, \bibinfo{person}{Michael~W Mahoney}, {and}
  \bibinfo{person}{Kurt Keutzer}.} \bibinfo{year}{2019}\natexlab{}.
\newblock \showarticletitle{{HAWQ}: Hessian aware quantization of neural
  networks with mixed-precision}. In \bibinfo{booktitle}{\emph{Proc. IEEE/CVF
  International Conference on Computer Vision}}. \bibinfo{pages}{293}.
\newblock
\showeprint{1905.03696}


\bibitem[Duarte et~al\mbox{.}(2018)]%
        {duarte2018fast}
\bibfield{author}{\bibinfo{person}{Javier Duarte} {et~al\mbox{.}}}
  \bibinfo{year}{2018}\natexlab{}.
\newblock \showarticletitle{Fast inference of deep neural networks in FPGAs for
  particle physics}.
\newblock \bibinfo{journal}{\emph{J. Instrum.}} \bibinfo{volume}{13},
  \bibinfo{number}{07} (\bibinfo{year}{2018}), \bibinfo{pages}{P07027}.
\newblock
\showeprint{1804.06913}


\bibitem[Fahim et~al\mbox{.}(2021)]%
        {fahim2021hls4ml}
\bibfield{author}{\bibinfo{person}{Farah Fahim} {et~al\mbox{.}}}
  \bibinfo{year}{2021}\natexlab{}.
\newblock \showarticletitle{hls4ml: An Open-Source Codesign Workflow to Empower
  Scientific Low-Power Machine Learning Devices}. In
  \bibinfo{booktitle}{\emph{1st TinyML Research Symposium}}.
\newblock
\showeprint[arxiv]{2103.05579}~[cs.LG]


\bibitem[Fu et~al\mbox{.}(2017)]%
        {fu2016deep}
\bibfield{author}{\bibinfo{person}{Yao Fu}, \bibinfo{person}{Ephrem Wu},
  \bibinfo{person}{Ashish Sirasao}, \bibinfo{person}{Sedny Attia},
  \bibinfo{person}{Kamran Khan}, {and} \bibinfo{person}{Ralph Wittig}.}
  \bibinfo{year}{2017}\natexlab{}.
\newblock \bibinfo{booktitle}{\emph{Deep learning with {INT8} optimization on
  {Xilinx} devices}}.
\newblock \bibinfo{type}{White Paper} WP486.
\newblock
\urldef\tempurl%
\url{https://docs.xilinx.com/v/u/en-US/wp486-deep-learning-int8}
\showURL{%
\tempurl}


\bibitem[Gholami et~al\mbox{.}(2021)]%
        {gholami2021survey}
\bibfield{author}{\bibinfo{person}{Amir Gholami}, \bibinfo{person}{Sehoon Kim},
  \bibinfo{person}{Zhen Dong}, \bibinfo{person}{Zhewei Yao},
  \bibinfo{person}{Michael~W Mahoney}, {and} \bibinfo{person}{Kurt Keutzer}.}
  \bibinfo{year}{2021}\natexlab{}.
\newblock \showarticletitle{A survey of quantization methods for efficient
  neural network inference}.
\newblock  (\bibinfo{year}{2021}).
\newblock
\showeprint{2103.13630}


\bibitem[Glorot and Bengio(2010)]%
        {glorot2010understanding}
\bibfield{author}{\bibinfo{person}{Xavier Glorot} {and} \bibinfo{person}{Yoshua
  Bengio}.} \bibinfo{year}{2010}\natexlab{}.
\newblock \showarticletitle{Understanding the difficulty of training deep
  feedforward neural networks}. In \bibinfo{booktitle}{\emph{Proc. 13th
  International Conference on Artificial Intelligence and Statistics}},
  \bibfield{editor}{\bibinfo{person}{Yee~Whye Teh} {and} \bibinfo{person}{Mike
  Titterington}} (Eds.), Vol.~\bibinfo{volume}{9}. \bibinfo{pages}{249}.
\newblock
\urldef\tempurl%
\url{https://proceedings.mlr.press/v9/glorot10a.html}
\showURL{%
\tempurl}


\bibitem[Glorot et~al\mbox{.}(2011)]%
        {relu2}
\bibfield{author}{\bibinfo{person}{Xavier Glorot}, \bibinfo{person}{Antoine
  Bordes}, {and} \bibinfo{person}{Yoshua Bengio}.}
  \bibinfo{year}{2011}\natexlab{}.
\newblock \showarticletitle{Deep Sparse Rectifier Neural Networks}. In
  \bibinfo{booktitle}{\emph{Proc. 14th International Conference on Artificial
  Intelligence and Statistics}}, \bibfield{editor}{\bibinfo{person}{Geoffrey
  Gordon}, \bibinfo{person}{David Dunson}, {and} \bibinfo{person}{Miroslav
  Dud\'{i}k}} (Eds.), Vol.~\bibinfo{volume}{15}. \bibinfo{pages}{315}.
\newblock
\urldef\tempurl%
\url{http://proceedings.mlr.press/v15/glorot11a.html}
\showURL{%
\tempurl}


\bibitem[He et~al\mbox{.}(2016a)]%
        {he_deep_2015}
\bibfield{author}{\bibinfo{person}{Kaiming He}, \bibinfo{person}{Xiangyu
  Zhang}, \bibinfo{person}{Shaoqing Ren}, {and} \bibinfo{person}{Jian Sun}.}
  \bibinfo{year}{2016}\natexlab{a}.
\newblock \showarticletitle{Deep {Residual} {Learning} for {Image}
  {Recognition}}. In \bibinfo{booktitle}{\emph{2016 IEEE Conference on Computer
  Vision and Pattern Recognition (CVPR)}}. \bibinfo{pages}{770}.
\newblock
\urldef\tempurl%
\url{https://doi.org/10.1109/CVPR.2016.90}
\showDOI{\tempurl}
\showeprint{1512.03385}


\bibitem[He et~al\mbox{.}(2016b)]%
        {he_resnet_2016}
\bibfield{author}{\bibinfo{person}{Kaiming He}, \bibinfo{person}{Xiangyu
  Zhang}, \bibinfo{person}{Shaoqing Ren}, {and} \bibinfo{person}{Jian Sun}.}
  \bibinfo{year}{2016}\natexlab{b}.
\newblock \showarticletitle{Identity Mappings in Deep Residual Networks}. In
  \bibinfo{booktitle}{\emph{ECCV 2016}},
  \bibfield{editor}{\bibinfo{person}{Bastian Leibe}, \bibinfo{person}{Jiri
  Matas}, \bibinfo{person}{Nicu Sebe}, {and} \bibinfo{person}{Max Welling}}
  (Eds.). \bibinfo{pages}{630}.
\newblock
\showISBNx{978-3-319-46493-0}
\urldef\tempurl%
\url{https://doi.org/10.1007/978-3-319-46493-0_38}
\showDOI{\tempurl}
\showeprint{1603.05027}


\bibitem[Hinton et~al\mbox{.}(2015)]%
        {hinton_distilling_2015}
\bibfield{author}{\bibinfo{person}{Geoffrey Hinton}, \bibinfo{person}{Oriol
  Vinyals}, {and} \bibinfo{person}{Jeff Dean}.}
  \bibinfo{year}{2015}\natexlab{}.
\newblock \showarticletitle{Distilling the {Knowledge} in a {Neural}
  {Network}}.
\newblock  (\bibinfo{year}{2015}).
\newblock
\showeprint{1503.02531}


\bibitem[Huang et~al\mbox{.}(2017)]%
        {huang2017densely}
\bibfield{author}{\bibinfo{person}{Gao Huang}, \bibinfo{person}{Zhuang Liu},
  \bibinfo{person}{Laurens Van Der~Maaten}, {and} \bibinfo{person}{Kilian~Q
  Weinberger}.} \bibinfo{year}{2017}\natexlab{}.
\newblock \showarticletitle{Densely connected convolutional networks}. In
  \bibinfo{booktitle}{\emph{Proc. IEEE conference on computer vision and
  pattern recognition}}. \bibinfo{pages}{4700}.
\newblock
\urldef\tempurl%
\url{https://doi.org/10.1109/CVPR.2017.243}
\showDOI{\tempurl}
\showeprint{1608.06993}


\bibitem[Jacob et~al\mbox{.}(2018)]%
        {jacob2018quantization}
\bibfield{author}{\bibinfo{person}{Benoit Jacob}, \bibinfo{person}{Skirmantas
  Kligys}, \bibinfo{person}{Bo Chen}, \bibinfo{person}{Menglong Zhu},
  \bibinfo{person}{Matthew Tang}, \bibinfo{person}{Andrew Howard},
  \bibinfo{person}{Hartwig Adam}, {and} \bibinfo{person}{Dmitry Kalenichenko}.}
  \bibinfo{year}{2018}\natexlab{}.
\newblock \showarticletitle{Quantization and training of neural networks for
  efficient integer-arithmetic-only inference}. In
  \bibinfo{booktitle}{\emph{Proc. IEEE conference on computer vision and
  pattern recognition}}. \bibinfo{pages}{2704}.
\newblock
\urldef\tempurl%
\url{https://doi.org/10.1109/CVPR.2018.00286}
\showDOI{\tempurl}


\bibitem[Juracy et~al\mbox{.}(2023)]%
        {juracy2023cnn}
\bibfield{author}{\bibinfo{person}{Leonardo~Rezende Juracy},
  \bibinfo{person}{Rafael Garibotti}, \bibinfo{person}{Fernando~Gehm Moraes},
  {et~al\mbox{.}}} \bibinfo{year}{2023}\natexlab{}.
\newblock \showarticletitle{From CNN to DNN Hardware Accelerators: A Survey on
  Design, Exploration, Simulation, and Frameworks}.
\newblock \bibinfo{journal}{\emph{Foundations and Trends{\textregistered} in
  Electronic Design Automation}} \bibinfo{volume}{13}, \bibinfo{number}{4}
  (\bibinfo{year}{2023}), \bibinfo{pages}{270--344}.
\newblock


\bibitem[Kriman et~al\mbox{.}(2020)]%
        {kriman2020quartznet}
\bibfield{author}{\bibinfo{person}{Samuel Kriman}, \bibinfo{person}{Stanislav
  Beliaev}, \bibinfo{person}{Boris Ginsburg}, \bibinfo{person}{Jocelyn Huang},
  \bibinfo{person}{Oleksii Kuchaiev}, \bibinfo{person}{Vitaly Lavrukhin},
  \bibinfo{person}{Ryan Leary}, \bibinfo{person}{Jason Li}, {and}
  \bibinfo{person}{Yang Zhang}.} \bibinfo{year}{2020}\natexlab{}.
\newblock \showarticletitle{{QuartzNet}: Deep automatic speech recognition with
  {1D} time-channel separable convolutions}. In \bibinfo{booktitle}{\emph{2020
  IEEE International Conference on Acoustics, Speech and Signal Processing
  (ICASSP)}}. \bibinfo{pages}{6124}.
\newblock
\urldef\tempurl%
\url{https://doi.org/10.1109/ICASSP40776.2020.9053889}
\showDOI{\tempurl}


\bibitem[Krizhevsky(2009)]%
        {cifardataset}
\bibfield{author}{\bibinfo{person}{Alex Krizhevsky}.}
  \bibinfo{year}{2009}\natexlab{}.
\newblock \showarticletitle{Learning Multiple Layers of Features from Tiny
  Images}.
\newblock \bibinfo{journal}{\emph{Tech Report}} (\bibinfo{year}{2009}).
\newblock


\bibitem[Li et~al\mbox{.}(2020)]%
        {li2020residual}
\bibfield{author}{\bibinfo{person}{Guilin Li}, \bibinfo{person}{Junlei Zhang},
  \bibinfo{person}{Yunhe Wang}, \bibinfo{person}{Chuanjian Liu},
  \bibinfo{person}{Matthias Tan}, \bibinfo{person}{Yunfeng Lin},
  \bibinfo{person}{Wei Zhang}, \bibinfo{person}{Jiashi Feng}, {and}
  \bibinfo{person}{Tong Zhang}.} \bibinfo{year}{2020}\natexlab{}.
\newblock \showarticletitle{Residual distillation: Towards portable deep neural
  networks without shortcuts}.
\newblock \bibinfo{journal}{\emph{Advances in Neural Information Processing
  Systems}}  \bibinfo{volume}{33} (\bibinfo{year}{2020}),
  \bibinfo{pages}{8935}.
\newblock
\urldef\tempurl%
\url{https://proceedings.neurips.cc/paper/2020/file/657b96f0592803e25a4f07166fff289a-Paper.pdf}
\showURL{%
\tempurl}


\bibitem[Li et~al\mbox{.}(2018)]%
        {li2018visualizing}
\bibfield{author}{\bibinfo{person}{Hao Li}, \bibinfo{person}{Zheng Xu},
  \bibinfo{person}{Gavin Taylor}, \bibinfo{person}{Christoph Studer}, {and}
  \bibinfo{person}{Tom Goldstein}.} \bibinfo{year}{2018}\natexlab{}.
\newblock \showarticletitle{Visualizing the loss landscape of neural nets}.
\newblock \bibinfo{journal}{\emph{Advances in neural information processing
  systems}}  \bibinfo{volume}{31} (\bibinfo{year}{2018}).
\newblock


\bibitem[Lin and Jegelka(2018)]%
        {lin2018resnet}
\bibfield{author}{\bibinfo{person}{Hongzhou Lin} {and}
  \bibinfo{person}{Stefanie Jegelka}.} \bibinfo{year}{2018}\natexlab{}.
\newblock \showarticletitle{Resnet with one-neuron hidden layers is a universal
  approximator}.
\newblock \bibinfo{journal}{\emph{Advances in neural information processing
  systems}}  \bibinfo{volume}{31} (\bibinfo{year}{2018}).
\newblock


\bibitem[Ma et~al\mbox{.}(2018)]%
        {ma2018shufflenet}
\bibfield{author}{\bibinfo{person}{Ningning Ma}, \bibinfo{person}{Xiangyu
  Zhang}, \bibinfo{person}{Hai-Tao Zheng}, {and} \bibinfo{person}{Jian Sun}.}
  \bibinfo{year}{2018}\natexlab{}.
\newblock \showarticletitle{{ShuffleNet} v2: Practical guidelines for efficient
  {CNN} architecture design}. In \bibinfo{booktitle}{\emph{Proc. European
  conference on computer vision (ECCV)}}. \bibinfo{pages}{116}.
\newblock
\urldef\tempurl%
\url{https://doi.org/10.1109/ASPCON49795.2020.9276669}
\showDOI{\tempurl}
\showeprint{1807.11164}


\bibitem[{Ma} et~al\mbox{.}(2018)]%
        {offchipmemory}
\bibfield{author}{\bibinfo{person}{Y. {Ma}}, \bibinfo{person}{Y. {Cao}},
  \bibinfo{person}{S. {Vrudhula}}, {and} \bibinfo{person}{J. {Seo}}.}
  \bibinfo{year}{2018}\natexlab{}.
\newblock \showarticletitle{Optimizing the Convolution Operation to Accelerate
  Deep Neural Networks on {FPGA}}.
\newblock \bibinfo{journal}{\emph{IEEE Trans Very Large Scale Integr. VLSI
  Syst.}} \bibinfo{volume}{26}, \bibinfo{number}{7} (\bibinfo{year}{2018}),
  \bibinfo{pages}{1354}.
\newblock
\urldef\tempurl%
\url{https://doi.org/10.1109/TVLSI.2018.2815603}
\showDOI{\tempurl}


\bibitem[{Ma} et~al\mbox{.}(2017)]%
        {controllogic17}
\bibfield{author}{\bibinfo{person}{Y. {Ma}}, \bibinfo{person}{M. {Kim}},
  \bibinfo{person}{Y. {Cao}}, \bibinfo{person}{S. {Vrudhula}}, {and}
  \bibinfo{person}{J. {Seo}}.} \bibinfo{year}{2017}\natexlab{}.
\newblock \showarticletitle{End-to-end scalable {FPGA} accelerator for deep
  residual networks}. In \bibinfo{booktitle}{\emph{2017 IEEE International
  Symposium on Circuits and Systems (ISCAS)}}. \bibinfo{pages}{1}.
\newblock
\urldef\tempurl%
\url{https://doi.org/10.1109/ISCAS.2017.8050344}
\showDOI{\tempurl}


\bibitem[Mirzadeh et~al\mbox{.}(2020)]%
        {Mirzadeh2020ImprovedKD}
\bibfield{author}{\bibinfo{person}{Seyed~Iman Mirzadeh},
  \bibinfo{person}{Mehrdad Farajtabar}, \bibinfo{person}{Ang Li},
  \bibinfo{person}{N. Levine}, \bibinfo{person}{A. Matsukawa}, {and}
  \bibinfo{person}{H. Ghasemzadeh}.} \bibinfo{year}{2020}\natexlab{}.
\newblock \showarticletitle{Improved Knowledge Distillation via Teacher
  Assistant}. In \bibinfo{booktitle}{\emph{AAAI}}, Vol.~\bibinfo{volume}{34}.
  \bibinfo{pages}{5191}.
\newblock
\urldef\tempurl%
\url{https://doi.org/10.1609/aaai.v34i04.5963}
\showDOI{\tempurl}
\showeprint{1902.03393}


\bibitem[Monti et~al\mbox{.}(2018)]%
        {avoidingskip18}
\bibfield{author}{\bibinfo{person}{Ricardo~Pio Monti}, \bibinfo{person}{Sina
  Tootoonian}, {and} \bibinfo{person}{Robin Cao}.}
  \bibinfo{year}{2018}\natexlab{}.
\newblock \showarticletitle{Avoiding Degradation in Deep Feed-Forward Networks
  by Phasing Out Skip-Connections}.
\newblock \bibinfo{journal}{\emph{Artificial Neural Networks and Machine
  Learning (ICANN)}}  \bibinfo{volume}{11141} (\bibinfo{year}{2018}).
\newblock
\urldef\tempurl%
\url{https://doi.org/10.1007/978-3-030-01424-7_44}
\showDOI{\tempurl}


\bibitem[Moons et~al\mbox{.}(2017)]%
        {moons2017minimum}
\bibfield{author}{\bibinfo{person}{Bert Moons}, \bibinfo{person}{Koen
  Goetschalckx}, \bibinfo{person}{Nick~Van Berckelaer}, {and}
  \bibinfo{person}{Marian Verhelst}.} \bibinfo{year}{2017}\natexlab{}.
\newblock \showarticletitle{Minimum Energy Quantized Neural Networks}. In
  \bibinfo{booktitle}{\emph{2017 51st Asilomar Conference on Signals, Systems,
  and Computers}}. \bibinfo{pages}{1921}.
\newblock
\urldef\tempurl%
\url{https://doi.org/10.1109/ACSSC.2017.8335699}
\showDOI{\tempurl}
\showeprint[arxiv]{1711.00215}~[cs.NE]


\bibitem[Nair and Hinton(2010)]%
        {relu1}
\bibfield{author}{\bibinfo{person}{Vinod Nair} {and}
  \bibinfo{person}{Geoffrey~E. Hinton}.} \bibinfo{year}{2010}\natexlab{}.
\newblock \showarticletitle{Rectified Linear Units Improve Restricted
  {Boltzmann} Machines}. In \bibinfo{booktitle}{\emph{Proc. 27th International
  Conference on Machine Learning}}. \bibinfo{pages}{807}.
\newblock
\urldef\tempurl%
\url{https://icml.cc/Conferences/2010/papers/432.pdf}
\showURL{%
\tempurl}


\bibitem[Netzer et~al\mbox{.}(2011)]%
        {svhndataset}
\bibfield{author}{\bibinfo{person}{Yuval Netzer}, \bibinfo{person}{Tao Wang},
  \bibinfo{person}{Adam Coates}, \bibinfo{person}{Alessandro Bissacco},
  \bibinfo{person}{Bo Wu}, {and} \bibinfo{person}{Andrew~Y. Ng}.}
  \bibinfo{year}{2011}\natexlab{}.
\newblock \showarticletitle{Reading Digits in Natural Images with Unsupervised
  Feature Learning}.
\newblock \bibinfo{journal}{\emph{NIPS Workshop on Deep Learning and
  Unsupervised Feature Learning}} (\bibinfo{year}{2011}).
\newblock


\bibitem[Pan and Yang(2009)]%
        {pan2009survey}
\bibfield{author}{\bibinfo{person}{Sinno~Jialin Pan} {and}
  \bibinfo{person}{Qiang Yang}.} \bibinfo{year}{2009}\natexlab{}.
\newblock \showarticletitle{A survey on transfer learning}.
\newblock \bibinfo{journal}{\emph{IEEE Trans. Knowl. Data Eng.}}
  \bibinfo{volume}{22}, \bibinfo{number}{10} (\bibinfo{year}{2009}),
  \bibinfo{pages}{1345}.
\newblock
\urldef\tempurl%
\url{https://doi.org/10.1109/TKDE.2009.191}
\showDOI{\tempurl}


\bibitem[Panayotov et~al\mbox{.}(2015)]%
        {panayotov2015librispeech}
\bibfield{author}{\bibinfo{person}{Vassil Panayotov}, \bibinfo{person}{Guoguo
  Chen}, \bibinfo{person}{Daniel Povey}, {and} \bibinfo{person}{Sanjeev
  Khudanpur}.} \bibinfo{year}{2015}\natexlab{}.
\newblock \showarticletitle{{LibriSpeech}: An {ASR} corpus based on public
  domain audio books}. In \bibinfo{booktitle}{\emph{2015 IEEE International
  Conference on Acoustics, Speech and Signal Processing (ICASSP)}}.
  \bibinfo{pages}{5206}.
\newblock
\urldef\tempurl%
\url{https://doi.org/10.1109/ICASSP.2015.7178964}
\showDOI{\tempurl}


\bibitem[Pappalardo(2022)]%
        {brevitas}
\bibfield{author}{\bibinfo{person}{Alessandro Pappalardo}.}
  \bibinfo{year}{2022}\natexlab{}.
\newblock \bibinfo{booktitle}{\emph{Xilinx/brevitas}}.
\newblock
\urldef\tempurl%
\url{https://doi.org/10.5281/zenodo.3333552}
\showDOI{\tempurl}


\bibitem[Paszke et~al\mbox{.}(2019)]%
        {NEURIPS2019_pytorch}
\bibfield{author}{\bibinfo{person}{Adam Paszke}, \bibinfo{person}{Sam Gross},
  \bibinfo{person}{Francisco Massa}, \bibinfo{person}{Adam Lerer},
  \bibinfo{person}{James Bradbury}, \bibinfo{person}{Gregory Chanan},
  \bibinfo{person}{Trevor Killeen}, \bibinfo{person}{Zeming Lin},
  \bibinfo{person}{Natalia Gimelshein}, \bibinfo{person}{Luca Antiga},
  \bibinfo{person}{Alban Desmaison}, \bibinfo{person}{Andreas Kopf},
  \bibinfo{person}{Edward Yang}, \bibinfo{person}{Zachary DeVito},
  \bibinfo{person}{Martin Raison}, \bibinfo{person}{Alykhan Tejani},
  \bibinfo{person}{Sasank Chilamkurthy}, \bibinfo{person}{Benoit Steiner},
  \bibinfo{person}{Lu Fang}, \bibinfo{person}{Junjie Bai}, {and}
  \bibinfo{person}{Soumith Chintala}.} \bibinfo{year}{2019}\natexlab{}.
\newblock \showarticletitle{{PyTorch}: An Imperative Style, High-Performance
  Deep Learning Library}.
\newblock In \bibinfo{booktitle}{\emph{{Advances in Neural Information
  Processing Systems}}}, \bibfield{editor}{\bibinfo{person}{H.~Wallach},
  \bibinfo{person}{H.~Larochelle}, \bibinfo{person}{A.~Beygelzimer},
  \bibinfo{person}{F.~d\textquotesingle Alch\'{e}-Buc},
  \bibinfo{person}{E.~Fox}, {and} \bibinfo{person}{R.~Garnett}} (Eds.).
  Vol.~\bibinfo{volume}{32}. \bibinfo{pages}{8024}.
\newblock
\showeprint{1912.01703}
\urldef\tempurl%
\url{http://papers.neurips.cc/paper/9015-pytorch-an-imperative-style-high-performance-deep-learning-library.pdf}
\showURL{%
\tempurl}


\bibitem[Petersen and Voigtlaender(2020)]%
        {petersen2020equivalence}
\bibfield{author}{\bibinfo{person}{Philipp Petersen} {and}
  \bibinfo{person}{Felix Voigtlaender}.} \bibinfo{year}{2020}\natexlab{}.
\newblock \showarticletitle{Equivalence of approximation by convolutional
  neural networks and fully-connected networks}.
\newblock \bibinfo{journal}{\emph{Proc. Amer. Math. Soc.}}
  \bibinfo{volume}{148}, \bibinfo{number}{4} (\bibinfo{year}{2020}),
  \bibinfo{pages}{1567--1581}.
\newblock


\bibitem[Sau and Balasubramanian(2016)]%
        {kd2016}
\bibfield{author}{\bibinfo{person}{Bharat~Bhusan Sau} {and}
  \bibinfo{person}{Vineeth~N. Balasubramanian}.}
  \bibinfo{year}{2016}\natexlab{}.
\newblock \showarticletitle{Deep Model Compression: Distilling Knowledge from
  Noisy Teachers}.
\newblock  (\bibinfo{year}{2016}).
\newblock
\showeprint[arxiv]{1610.09650}


\bibitem[Silvestre-Ryan and Holmes(2021)]%
        {silvestre2021pair}
\bibfield{author}{\bibinfo{person}{Jordi Silvestre-Ryan} {and}
  \bibinfo{person}{Ian Holmes}.} \bibinfo{year}{2021}\natexlab{}.
\newblock \showarticletitle{Pair consensus decoding improves accuracy of neural
  network basecallers for nanopore sequencing}.
\newblock \bibinfo{journal}{\emph{Genome Biol.}}  \bibinfo{volume}{22}
  (\bibinfo{year}{2021}), \bibinfo{pages}{38}.
\newblock
\urldef\tempurl%
\url{https://doi.org/10.1186/s13059-020-02255-1}
\showDOI{\tempurl}


\bibitem[Simonyan and Zisserman(2015)]%
        {vgg2015}
\bibfield{author}{\bibinfo{person}{Karen Simonyan} {and}
  \bibinfo{person}{Andrew Zisserman}.} \bibinfo{year}{2015}\natexlab{}.
\newblock \showarticletitle{Very Deep Convolutional Networks for Large-Scale
  Image Recognition}. In \bibinfo{booktitle}{\emph{3rd International Conference
  on Learning Representations, {ICLR} 2015, San Diego, CA, USA, May 7-9, 2015,
  Conference Track Proceedings}}, \bibfield{editor}{\bibinfo{person}{Yoshua
  Bengio} {and} \bibinfo{person}{Yann LeCun}} (Eds.).
\newblock
\urldef\tempurl%
\url{http://arxiv.org/abs/1409.1556}
\showURL{%
\tempurl}


\bibitem[Tarvainen and Valpola(2017)]%
        {kd2017}
\bibfield{author}{\bibinfo{person}{Antti Tarvainen} {and}
  \bibinfo{person}{Harri Valpola}.} \bibinfo{year}{2017}\natexlab{}.
\newblock \showarticletitle{Mean Teachers Are Better Role Models:
  Weight-Averaged Consistency Targets Improve Semi-Supervised Deep Learning
  Results}. In \bibinfo{booktitle}{\emph{Advances in Neural Information
  Processing Systems}}, \bibfield{editor}{\bibinfo{person}{I.~Guyon},
  \bibinfo{person}{U.~Von Luxburg}, \bibinfo{person}{S.~Bengio},
  \bibinfo{person}{H.~Wallach}, \bibinfo{person}{R.~Fergus},
  \bibinfo{person}{S.~Vishwanathan}, {and} \bibinfo{person}{R.~Garnett}}
  (Eds.), Vol.~\bibinfo{volume}{30}. \bibinfo{pages}{1195}.
\newblock
\showISBNx{9781510860964}
\showeprint{1703.01780}
\urldef\tempurl%
\url{https://proceedings.neurips.cc/paper/2017/file/68053af2923e00204c3ca7c6a3150cf7-Paper.pdf}
\showURL{%
\tempurl}


\bibitem[Veit et~al\mbox{.}(2016)]%
        {veit2016residual}
\bibfield{author}{\bibinfo{person}{Andreas Veit}, \bibinfo{person}{Michael~J
  Wilber}, {and} \bibinfo{person}{Serge Belongie}.}
  \bibinfo{year}{2016}\natexlab{}.
\newblock \showarticletitle{Residual networks behave like ensembles of
  relatively shallow networks}.
\newblock \bibinfo{journal}{\emph{Advances in Neural Information Processing
  Systems}}  \bibinfo{volume}{29} (\bibinfo{year}{2016}).
\newblock
\showeprint{1605.06431}
\urldef\tempurl%
\url{https://proceedings.neurips.cc/paper/2016/file/37bc2f75bf1bcfe8450a1a41c200364c-Paper.pdf}
\showURL{%
\tempurl}


\bibitem[Wang et~al\mbox{.}(2019)]%
        {wang2019haq}
\bibfield{author}{\bibinfo{person}{Kuan Wang}, \bibinfo{person}{Zhijian Liu},
  \bibinfo{person}{Yujun Lin}, \bibinfo{person}{Ji Lin}, {and}
  \bibinfo{person}{Song Han}.} \bibinfo{year}{2019}\natexlab{}.
\newblock \showarticletitle{{HAQ}: Hardware-aware automated quantization with
  mixed precision}. In \bibinfo{booktitle}{\emph{Proc. IEEE/CVF Conference on
  Computer Vision and Pattern Recognition}}. \bibinfo{pages}{8612}.
\newblock
\urldef\tempurl%
\url{https://doi.org/10.1109/CVPR.2019.00881}
\showDOI{\tempurl}
\showeprint{1811.08886}


\bibitem[Weiss et~al\mbox{.}(2016)]%
        {weiss2016survey}
\bibfield{author}{\bibinfo{person}{Karl Weiss}, \bibinfo{person}{Taghi~M
  Khoshgoftaar}, {and} \bibinfo{person}{DingDing Wang}.}
  \bibinfo{year}{2016}\natexlab{}.
\newblock \showarticletitle{A survey of transfer learning}.
\newblock \bibinfo{journal}{\emph{J. Big Data}} \bibinfo{volume}{3},
  \bibinfo{number}{1} (\bibinfo{year}{2016}), \bibinfo{pages}{1}.
\newblock
\urldef\tempurl%
\url{https://doi.org/10.1186/s40537-016-0043-6}
\showDOI{\tempurl}


\bibitem[Yang et~al\mbox{.}(2019)]%
        {yang2019synetgy}
\bibfield{author}{\bibinfo{person}{Yifan Yang}, \bibinfo{person}{Qijing Huang},
  \bibinfo{person}{Bichen Wu}, \bibinfo{person}{Tianjun Zhang},
  \bibinfo{person}{Liang Ma}, \bibinfo{person}{Giulio Gambardella},
  \bibinfo{person}{Michaela Blott}, \bibinfo{person}{Luciano Lavagno},
  \bibinfo{person}{Kees Vissers}, \bibinfo{person}{John Wawrzynek},
  {et~al\mbox{.}}} \bibinfo{year}{2019}\natexlab{}.
\newblock \showarticletitle{Synetgy: Algorithm-hardware co-design for convnet
  accelerators on embedded {FPGAs}}. In \bibinfo{booktitle}{\emph{Proc. 2019
  ACM/SIGDA International Symposium on Field-Programmable Gate Arrays}}.
  \bibinfo{pages}{23}.
\newblock
\urldef\tempurl%
\url{https://doi.org/10.1145/3289602.3293902}
\showDOI{\tempurl}


\bibitem[Yao et~al\mbox{.}(2020)]%
        {yao2020pyhessian}
\bibfield{author}{\bibinfo{person}{Zhewei Yao}, \bibinfo{person}{Amir Gholami},
  \bibinfo{person}{Kurt Keutzer}, {and} \bibinfo{person}{Michael~W Mahoney}.}
  \bibinfo{year}{2020}\natexlab{}.
\newblock \showarticletitle{Pyhessian: Neural networks through the lens of the
  hessian}. In \bibinfo{booktitle}{\emph{2020 IEEE international conference on
  big data (Big data)}}. IEEE, \bibinfo{pages}{581--590}.
\newblock


\bibitem[Zagoruyko and Komodakis(2017)]%
        {diracnets17}
\bibfield{author}{\bibinfo{person}{Sergey Zagoruyko} {and}
  \bibinfo{person}{Nikos Komodakis}.} \bibinfo{year}{2017}\natexlab{}.
\newblock \showarticletitle{{DiracNets}: Training Very Deep Neural Networks
  Without Skip-Connections}.
\newblock  (\bibinfo{year}{2017}).
\newblock
\showeprint[arxiv]{1706.00388}


\bibitem[Zagoruyko and Komodakis(2018)]%
        {zagoruyko_diracnets_2018}
\bibfield{author}{\bibinfo{person}{Sergey Zagoruyko} {and}
  \bibinfo{person}{Nikos Komodakis}.} \bibinfo{year}{2018}\natexlab{}.
\newblock \showarticletitle{{DiracNets}: {Training} {Very} {Deep} {Neural}
  {Networks} {Without} {Skip}-{Connections}}.
\newblock  (\bibinfo{year}{2018}).
\newblock
\showeprint[arxiv]{1706.00388}


\bibitem[Zhuang et~al\mbox{.}(2020)]%
        {zhuang2020comprehensive}
\bibfield{author}{\bibinfo{person}{Fuzhen Zhuang}, \bibinfo{person}{Zhiyuan
  Qi}, \bibinfo{person}{Keyu Duan}, \bibinfo{person}{Dongbo Xi},
  \bibinfo{person}{Yongchun Zhu}, \bibinfo{person}{Hengshu Zhu},
  \bibinfo{person}{Hui Xiong}, {and} \bibinfo{person}{Qing He}.}
  \bibinfo{year}{2020}\natexlab{}.
\newblock \showarticletitle{A comprehensive survey on transfer learning}.
\newblock \bibinfo{journal}{\emph{Proc. IEEE}} \bibinfo{volume}{109},
  \bibinfo{number}{1} (\bibinfo{year}{2020}), \bibinfo{pages}{43}.
\newblock
\urldef\tempurl%
\url{https://doi.org/10.1109/JPROC.2020.3004555}
\showDOI{\tempurl}


\end{thebibliography}

\end{document}